\documentclass{seekjudge}

\usepackage{amssymb}
\usepackage{colortbl}
\usepackage{array}
\usepackage{adjustbox}
\usepackage{xspace}
\usepackage{enumitem}
\usepackage{fontawesome5}
\usepackage{makecell}
\usepackage{placeins}

\usepackage{tikz}
\usetikzlibrary{positioning,arrows.meta,shapes.geometric,fit,backgrounds,calc,decorations.pathreplacing}
\usepackage{amsmath}
\usepackage{tabularx}

\definecolor{prcA}{RGB}{244,245,247}   %
\definecolor{prcB}{RGB}{226,236,248}   %
\definecolor{prcC}{RGB}{198,219,239}   %
\definecolor{prcD}{RGB}{148,186,225}   %
\definecolor{prcE}{RGB}{74,133,193}    %
\definecolor{prcF}{RGB}{31,78,156}     %

\newcommand{\prn}[1]{{\scriptsize\leavevmode #1}}             %
\newcommand{\prw}[1]{{\scriptsize\hypersetup{citecolor=white}\textcolor{white}{#1}}} %

\newcommand{\heatbar}{%
\begin{tikzpicture}[baseline=(current bounding box.center)]
  \foreach \hc/\hl [count=\hi from 0] in {prcA/none,prcB/1,prcC/2--3,prcD/4--6,prcE/7--11,prcF/12+}{%
    \fill[\hc] (0,0.42*\hi) rectangle (0.30,0.42*\hi+0.42);
    \node[anchor=west,font=\scriptsize,inner sep=1.2pt] at (0.31,0.42*\hi+0.21) {\hl};
  }
  \draw[black!30,line width=0.3pt] (0,0) rectangle (0.30,2.52);
\end{tikzpicture}%
}

\definecolor{grpgreen}{RGB}{223,240,216}
\definecolor{grpyellow}{RGB}{252,248,227}
\definecolor{grpred}{RGB}{242,222,222}
\definecolor{cellhot}{RGB}{198,219,239}
\definecolor{cellwarm}{RGB}{222,235,247}
\definecolor{cellempty}{RGB}{252,224,224}

\graphicspath{{figures/}}

\title{%
\begin{center}
No Free Checker: \\
A Survey of Verifiers for Robot Policies
\end{center}
}
\renewcommand{\headerbrand}{Preprint}

\affiliation{Zhejiang University}
\affiliation{City University of Hong Kong}
\author[\affil{Zhejiang University}]{Yang Wan}
\author[\affil{Zhejiang University}]{Xihang Yue}
\author[\affil{Zhejiang University}]{Zhirui Liu}
\author[\affil{Zhejiang University}]{Ziyuan Chu}
\author[\affil{Zhejiang University}]{Shuxun Wang}
\author[\affil{Zhejiang University}]{Yuhan Chen}
\author[\affil{City University of Hong Kong}]{Xiaonan Jiang}
\author[\affil{Zhejiang University}]{Xukun Zhu}
\author[\affil{Zhejiang University}]{Yubo Dong}
\author[\affil{Zhejiang University}]{Linchao Zhu}

\newcommand{\ghlink}{https://github.com/ZJUSCL/Awesome-Robot-Verifier}
\newcommand{\ghname}{Awesome-Robot-Verifier}
\coverlinks{{\color{metafg}\faGithub~\href{\ghlink}{\textcolor{metablue}{\ghname}}}}

\abstract{A verifier for robot policies reads a candidate behavior and returns a score for how well it did, used both to evaluate vision-language-action policies and to train them.
Verifiers range from success detectors and reward models to runtime monitors, safety filters, and temporal-logic specifications.

We survey roughly 150 verifiers and compare them along two properties.
\textit{Availability} is how much a verdict costs, how early in a rollout the verdict arrives, and how often a verdict can be asked for. Availability rises as verdicts get cheaper, earlier, and denser.
\textit{Credibility} is how much a high score tells us about the task. Credibility falls as the judgment becomes gameable and self-serving.
We group the verifiers by who supplies the judgment: \textit{human verifiers}, \textit{rule-based and formal verifiers}, \textit{learned and pretrained verifiers}, and \textit{model-intrinsic verifiers}.
Across the four families, we find that credibility falls as availability rises. Regardless of who supplies the judgment, \textbf{there is no free checker}.

We then examine what validates a verifier itself, and how much a high score tells us.
Three measures appear in the literature:
agreement with human labels, the performance of the policy it trains, and behavior under reward hacking.
We close with nine metrics that make a verifier claim checkable, and coordinates for the verifiers still to be built.
}

\metadata[Keywords]{Robot policy evaluation, Reward models, Vision-language-action models, Runtime monitoring, Safety filters, Reward hacking, World models, Data curation for imitation learning}

\newcommand{\cmark}{\textcolor{green!55!black}{\checkmark}}
\newcommand{\xmark}{\textcolor{red!70!black}{\ensuremath{\times}}}

\begin{document}

\makeatletter
\gdef\seek@titleimage{%
  \vspace{0.3cm}%
  \includegraphics[width=\linewidth]{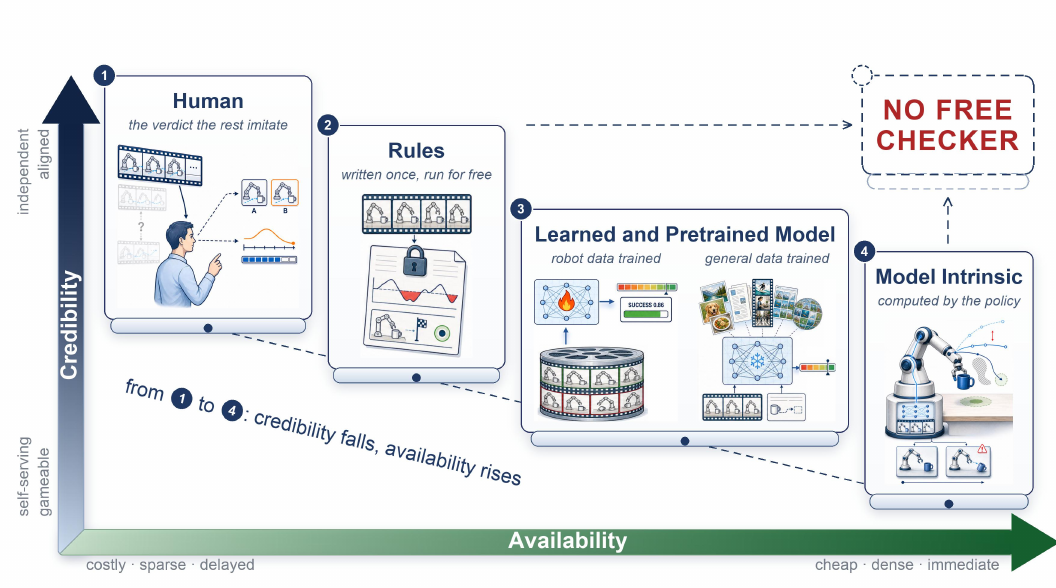}%
  \vspace{0.3cm}%
}
\makeatother

\maketitle

\seektoc

\section{Introduction}
\label{sec:intro}

In robot learning, vision-language-action (VLA) policies~\citep{brohan2023rt2,kim2024openvla,black2025pi0,PI05,pi07_2026} and robot world models~\citep{Cosmos,deepmind2025genie3} are scaling rapidly and becoming more capable.
While scaling remains important, verification of training data and model behavior is also essential, as demonstrated by recent progress in large language models~\citep{DeepSeek-R1}.
Such verification signals are used at multiple stages of robot learning, including filtering and reweighting demonstrations for pre-training~\citep{Re-Mix,Demo-SCORE,DataMIL}, providing reward signals during policy post-training~\citep{SimpleVLA-RL,RIPT-VLA,RLinf-VLA}, ranking candidate actions at inference time~\citep{RoboMonkey,V-GPS,MG-Select}, and evaluating policy performance using rollouts generated by world models~\citep{WorldEval,WorldGym,GigaWorld-1}.
We refer to the mechanisms that provide these verification signals collectively as \emph{verifiers}.

A robot verifier is a mapping
\begin{equation}
\text{verifier}:\ (\text{context},\, \text{candidate}) \;\longmapsto\; \text{score}.
\label{eq:verifier}
\end{equation}
The context comprises observations, a goal, and possibly a language instruction. The score can be boolean, scalar, vector-valued, or a distribution. The candidate varies the most of the three, running from a single action chunk up to an entire policy, with subtask segments and full trajectories in between.

Building reliable \emph{verifiers} for robot learning is particularly difficult because the physical world lacks the cheap and well-defined criteria available in mathematics and code.
First, determining whether a physical task has been completed correctly is costly because task success is often inferred from imperfect sensory observations~\citep{Robometer,RoboReward,Robo-Dopamine}.
Second, evaluating a policy is expensive because testing a policy may require real-robot rollouts, which consume physical time and limit large-scale sampling~\citep{WorldEval,GigaWorld-1,InteractiveWS}.
Third, verification often has to happen before the task is finished, because a single incorrect action is enough to cause task failure or hardware damage~\citep{Code-as-Monitor,SAFE,Sentinel,PNCBF,aegis2026}.
Finally, task performance is often continuous, as robot behaviors can differ in progress, safety, and execution quality, making a single success-or-failure label insufficient~\citep{GVL,VIP,LIV,TOPReward}.

\begin{figure}[!t]
\centering
\definecolor{tlnavy}{RGB}{27,48,84}
\definecolor{tlgrey}{RGB}{130,134,142}

{\scriptsize
\colorlet{tlcitegrey}{tlnavy!62}
\hypersetup{citecolor=tlcitegrey}
\newcommand{\tlcite}[1]{{\tiny\citep{#1}}}
\begin{tikzpicture}[x=0.1130\linewidth,y=1.35cm,
                    every node/.style={inner sep=1pt,outer sep=0pt}]

\foreach \sx/\yr in {0/2020,1/2021,2/2022,3/2023,4/2024,5/2025,6/2026} {
  \draw[black!12,line width=0.4pt] (\sx,0.40) -- (\sx,5.90);
  \node[anchor=north,font=\small\color{tlgrey}] at (\sx,0.30) {\yr};
}
\draw[-{Stealth[length=4pt,width=3pt]},tlgrey,line width=0.5pt]
  (-0.42,0.40) -- (7.10,0.40);

\foreach \ly/\lname in {5/{Human}, 4/{Rules and formal}, 3/{Trained model},
                        2/{Training-free\\model}, 1/{Model-intrinsic}} {
  \node[anchor=east,align=right,font=\small\bfseries\color{tlnavy}] at (-0.30,\ly) {\lname};
}
\foreach \ly/\lb/\le in {5/0/6, 4/0/6, 3/2/6, 2/3/6, 1/0/6} {
  \draw[tlnavy,line width=1.1pt] (\lb,\ly) -- (\le,\ly);
}

\foreach \sx/\ly/\n/\ne/\lab/\col in {%
  0/5/1/1/{RewardSketch~\tlcite{RewardSketching}}/tlnavy,
  1/5/2/2/{PEBBLE~\tlcite{PEBBLE}\\B-Pref~\tlcite{BPref}}/tlnavy,
  2/5/2/2/{HACO~\tlcite{HACO}}/tlnavy,
  3/5/1/1/{Sirius~\tlcite{Sirius}}/tlnavy,
  4/5/1/1/{Sirius-Fleet~\tlcite{liu2024siriusfleet}}/tlnavy,
  5/5/4/4/{RoboArena~\tlcite{RoboArena}\\HIL-SERL~\tlcite{HIL-SERL}\\VeoEval~\tlcite{VeoEval}}/tlnavy,
  0/4/1/1/{RTD~\tlcite{RTD}}/tlnavy,
  1/4/3/3/{RTS~\tlcite{RTS}\\pred. filter~\tlcite{wabersich2021psf}}/tlnavy,
  2/4/5/5/{SaRA~\tlcite{SaRA}\\BRSL~\tlcite{BRSL}\\CALVIN~\tlcite{CALVIN}}/tlnavy,
  3/4/5/5/{conf. STL~\tlcite{ConformalSTL}\\LIBERO~\tlcite{LIBERO}\\MimicGen~\tlcite{MimicGen}}/tlnavy,
  4/4/14/14/{Eureka~\tlcite{Eureka}\\PNCBF~\tlcite{PNCBF}\\ReKep~\tlcite{ReKep}}/tlnavy,
  5/4/12/12/{CodeMonitor~\tlcite{Code-as-Monitor}\\TGPO~\tlcite{TGPO}\\ARMOUR~\tlcite{ARMOUR}}/tlnavy,
  6/4/6/12/{SafeManip~\tlcite{SafeManip}\\RINSE~\tlcite{RINSE}\\SimpleVLA-RL~\tlcite{SimpleVLA-RL}}/tlnavy,
  2/3/3/3/{DWBC~\tlcite{DWBC}\\ILEED~\tlcite{ILEED}}/tlnavy,
  3/3/5/5/{VIP~\tlcite{VIP}\\LIV~\tlcite{LIV}\\SayCan~\tlcite{SayCan}}/tlnavy,
  4/3/2/2/{RoboCat~\tlcite{RoboCat}}/tlnavy,
  5/3/18/18/{RECAP~\tlcite{RECAP}\\VLAC~\tlcite{VLAC}}/tlnavy,
  6/3/15/30/{Robometer~\tlcite{Robometer}\\DataMIL~\tlcite{DataMIL}\\GigaWorld-1~\tlcite{GigaWorld-1}}/tlnavy,
  3/2/3/3/{KnowNo~\tlcite{KnowNo}\\REFLECT~\tlcite{liu2023reflect}}/tlnavy,
  4/2/2/2/{VLM-RM~\tlcite{rocamonde2024vlmrm}\\GRAPE~\tlcite{GRAPE}}/tlnavy,
  5/2/3/3/{GVL~\tlcite{GVL}\\WorldEval~\tlcite{WorldEval}}/tlnavy,
  6/2/3/6/{TOPReward~\tlcite{TOPReward}\\SRPO~\tlcite{SRPO}}/tlnavy,
  0/1/2/2/{MOPO~\tlcite{MOPO}\\MOReL~\tlcite{MOReL}}/tlnavy,
  1/1/1/1/{LOMPO~\tlcite{LOMPO}}/tlnavy,
  2/1/1/1/{LS3~\tlcite{LS3}}/tlnavy,
  3/1/1/1/{VIPER~\tlcite{VIPER}}/tlnavy,
  4/1/1/1/{DiffReward~\tlcite{DiffusionReward}}/tlnavy,
  5/1/11/11/{SAFE~\tlcite{SAFE}\\Sentinel~\tlcite{Sentinel}\\TACO~\tlcite{TACO}}/tlnavy,
  6/1/6/12/{MG-Select~\tlcite{MG-Select}\\HideSeek~\tlcite{HideAndSeek}\\AnySafe~\tlcite{agrawal2025anysafe}}/tlnavy}
{
  \pgfmathsetmacro{\rr}{0.25 + 1.55*sqrt(\n)}
  \pgfmathsetmacro{\re}{0.25 + 1.55*sqrt(\ne)}
  \pgfmathsetmacro{\xoff}{(0.78*\re + 2.2)/51.0}
  \pgfmathsetmacro{\yoff}{(0.78*\re + 2.2)/33.0}
  \ifdim\n pt<\ne pt
    \draw[\col,line width=0.5pt,dashed,dash pattern=on 1.2pt off 1.2pt]
      (\sx,\ly) circle (\re pt);
  \fi
  \fill[\col] (\sx,\ly) circle (\rr pt);
  \node[anchor=south west,align=left,font=\scriptsize\color{tlnavy},inner sep=0pt]
    at (\sx+\xoff,\ly+\yoff) {\lab};
}

\node[anchor=north,align=center,font=\scriptsize\color{tlgrey}] at (6.20,-0.06)
  {dashed ring: full-year estimate};

\end{tikzpicture}
\par}
\caption{Representative systems by judge source and year, using the same five columns as Table~\ref{tab:property}. Disc area is the number of papers we cover in the corresponding year. Our search closes in early July 2026, so each 2026 disc carries a dashed ring at the full-year estimate, twice the observed count at the January-to-June rate.}
\label{fig:timeline}
\end{figure}
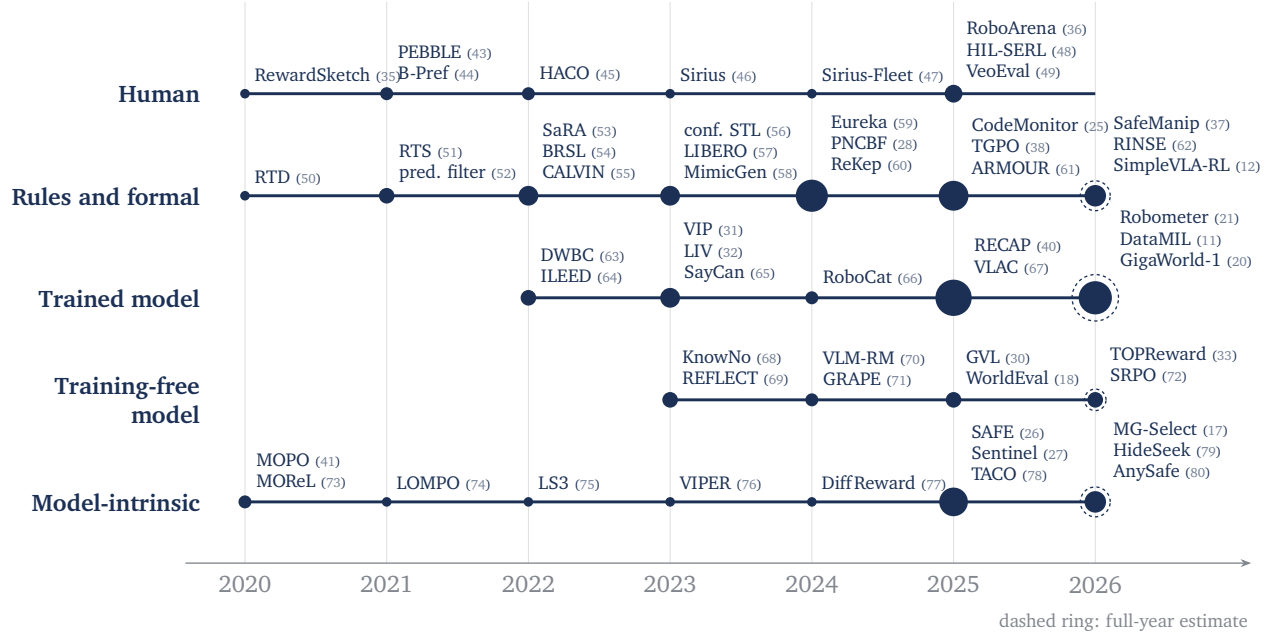

Given these challenges, different verifiers offer different strengths and limitations in robot learning.
To systematically understand existing methods, we survey roughly 150 works on verification in robot learning.
Figure~\ref{fig:timeline} places representative systems by family and year, and Table~\ref{tab:property} records what each family's score asserts.
Further, we characterize these works along two properties, \emph{availability} and \emph{credibility}.
\emph{Availability} is how much a verdict costs, how early in a rollout the verdict arrives, and how often a verdict can be asked for.
\emph{Credibility} is how much a high score tells us about the task.

The source of judgment strongly affects the \emph{availability} and \emph{credibility} of a verifier. We therefore group existing methods by judge source into four families: (a) \textbf{human verifiers}, where the judgment is provided directly by a person, (b) \textbf{rule-based and formal verifiers}, where the judgment is determined by predefined criteria, (c) \textbf{learned and pretrained verifiers}, where the judgment is produced by a neural model, (d) \textbf{model-intrinsic verifiers}, where the judgment comes from signals already computed by the model itself.

\begin{itemize}[leftmargin=1.5em]
\item \textbf{Human verifiers} (Section~\ref{sec:human}).
A person directly judges task success, compares trajectories, or intervenes during execution~\citep{ChristianoPrefs,RewardSketching,RoboArena}.
In terms of \emph{availability}, human judgments are costly and sparse because judgments take human effort.
In terms of \emph{credibility}, they provide a direct reference to task intent, although they may still be subjective or inconsistent.
\item \textbf{Rule-based and formal verifiers} (Section~\ref{sec:rules}).
The judgment is determined by predefined criteria, such as task predicates, temporal-logic specifications, or formal safety conditions~\citep{SafeManip,TGPO,PNCBF}.
In terms of \emph{availability}, these verifiers are inexpensive and repeatable once the required state information is available.
In terms of \emph{credibility}, they can provide strong and sometimes formal guarantees, but only when the predefined criteria, state estimates, and dynamics assumptions accurately represent the task.
\item \textbf{Learned and pretrained verifiers} (Section~\ref{sec:scorers}).
A neural model produces the judgment, either after task-specific training or directly from pretraining~\citep{SuccessVQA,RECAP,TOPReward}.
In terms of \emph{availability}, these verifiers are inexpensive to query and can provide dense feedback across many tasks and trajectories.
In terms of \emph{credibility}, their judgments depend on model accuracy, calibration, and generalization beyond the data on which they were trained or validated.
\item \textbf{Model-intrinsic verifiers} (Section~\ref{sec:intrinsic}).
The judgment is derived from signals already computed by the policy or predictive model, such as action uncertainty, model uncertainty, or learned reachability~\citep{MG-Select,MOPO,LatentSafetyFilters}.
In terms of \emph{availability}, these signals are the easiest to obtain because they require little or no additional computation or external supervision.
In terms of \emph{credibility}, model-intrinsic signals describe the model itself, and their relation to actual task performance is therefore indirect.
\end{itemize}

\begin{table}[!t]
\centering

\newcommand{\prgroup}[1]{\smash{\itshape\color{black!62}#1}}

\newcolumntype{P}{>{\centering\arraybackslash}m{2.12cm}}
\newcolumntype{L}{>{\raggedright\arraybackslash}m{3.45cm}}

{\footnotesize
\setlength{\tabcolsep}{3pt}
\setlength{\aboverulesep}{0pt}
\begin{tabular}{@{}c@{\hspace{7pt}}c@{}}
\begin{NiceTabular}{@{}L PPPPP@{}}[cell-space-top-limit=3pt,cell-space-bottom-limit=3.5pt]
\CodeBefore
  \cellcolor{prcF}{4-3,4-4,6-3,10-6}
  \cellcolor{prcE}{5-4,6-6,12-4}
  \cellcolor{prcD}{4-5,5-5,8-2,10-2,10-4}
  \cellcolor{prcC}{4-2,5-6,6-2,7-3,7-4,10-5}
  \cellcolor{prcB}{4-6,5-2,6-4,6-5,7-5,8-4,10-3,12-2,12-3}
  \cellcolor{prcA}{5-3,7-2,7-6,8-3,8-5,8-6,12-5,12-6}
\Body
\toprule
 & & & \multicolumn{2}{c}{\textbf{Learned and pretrained verifiers}} & \\
\cmidrule(lr){4-5}
\textbf{What the score asserts} & \textbf{Human} & \textbf{Rules and formal} & \textbf{Trained model} & \textbf{Training-free model} & \textbf{Model-intrinsic} \\
\midrule
\multicolumn{6}{@{}l}{\prgroup{a claim about the behavior}} \\
\quad task success
  & \prn{HIL-SERL~\citep{HIL-SERL}}
  & \prw{MimicGen~\citep{MimicGen}, Eureka~\citep{Eureka}, SimpleVLA-RL~\citep{SimpleVLA-RL}}
  & \prw{RECAP~\citep{RECAP}, RoboMonkey~\citep{RoboMonkey}, SuccessVQA~\citep{SuccessVQA}}
  & \prn{WorldEval~\citep{WorldEval}, WorldGym~\citep{WorldGym}}
  & \prn{MG-Select~\citep{MG-Select}} \\
\quad progress
  & \prn{reward sketching~\citep{RewardSketching}}
  &
  & \prw{VIP~\citep{VIP}, VLAC~\citep{VLAC}, Robo-Dopamine~\citep{Robo-Dopamine}}
  & \prn{GVL~\citep{GVL}, TOPReward~\citep{TOPReward}}
  & \prn{VIPER~\citep{VIPER}, Sentinel~\citep{Sentinel}} \\
\quad safety, constraint
  & \prn{ThriftyDAgger~\citep{ThriftyDAgger}, HG-DAgger~\citep{HGDagger}}
  & \prw{PNCBF~\citep{PNCBF}, RTD~\citep{RTD}, SafeManip~\citep{SafeManip}, ConformalSTL~\citep{ConformalSTL}}
  & \prn{RealTimeVerif~\citep{RealTimeVerif}}
  & \prn{GRAPE~\citep{GRAPE}}
  & \prw{LS3~\citep{LS3}, latent safety filters~\citep{LatentSafetyFilters}} \\
\quad execution quality
  &
  & \prn{RINSE~\citep{RINSE}, Video2Reward~\citep{Video2Reward}}
  & \prn{RECAP~\citep{RECAP}, Robo-Dopamine~\citep{Robo-Dopamine}}
  & \prn{GRAPE~\citep{GRAPE}}
  & \\
\quad human preference
  & \prn{PEBBLE~\citep{PEBBLE}, RoboArena~\citep{RoboArena}, B-Pref~\citep{BPref}}
  &
  & \prn{Robometer~\citep{Robometer}}
  &
  & \\
\midrule
\multicolumn{6}{@{}l}{\prgroup{a claim about the policy}} \\
\quad failure, uncertainty
  & \prn{ThriftyDAgger~\citep{ThriftyDAgger}, Sirius~\citep{Sirius}}
  & \prn{Code-as-Monitor~\citep{Code-as-Monitor}}
  & \prn{Foresight~\citep{Foresight}, AHA~\citep{AHA}, RoboFAC~\citep{RoboFAC}}
  & \prn{KnowNo~\citep{KnowNo}, FOREWARN~\citep{FOREWARN}}
  & \prw{SAFE~\citep{SAFE}, Sentinel~\citep{Sentinel}} \\
\midrule
\multicolumn{6}{@{}l}{\prgroup{a claim about the training data}} \\
\quad training value of a demonstration
  & \prn{Sirius~\citep{Sirius}}
  & \prn{RINSE~\citep{RINSE}}
  & \prw{Re-Mix~\citep{Re-Mix}, DataMIL~\citep{DataMIL}, Demo-SCORE~\citep{Demo-SCORE}}
  &
  & \\
\bottomrule
\end{NiceTabular}
&
\heatbar
\\
\end{tabular}
\par}

\caption{What the score asserts, plotted against the judge source supplying the criterion. Each cell names representative systems, and the fill depth is the number of papers there. Each paper is counted in exactly one judge-source column, and a paper asserting two properties appears in both rows.}
\label{tab:property}
\end{table}

Across these four families, the source of judgment shows a clear trade-off between \emph{availability} and \emph{credibility}.
Human judgments and explicit task criteria provide a more direct measure of task performance, but human judgment is costly to collect at scale, and an explicit criterion holds only where its state estimates and physical assumptions represent the task~\citep{RoboArena,SafeManip,PNCBF}.
Learned and model-intrinsic signals are cheaper to query and can provide denser feedback, but a learned judgment depends on how well the model generalizes beyond its training data, and an intrinsic signal describes the model itself rather than task performance~\citep{Robometer,TOPReward,MG-Select,MOPO}.
Therefore, in the physical world, obtaining both high \emph{availability} and high \emph{credibility} at the same time is costly.
We refer to this observation as \textbf{no free checker}.

We therefore examine in Section~\ref{sec:meta} how a verifier's own error is measured, and why \emph{credibility} is difficult to establish in robotics.
Three metrics appear, agreement with a fixed reference, the policy that results from training on the verifier, and the verifier's behavior under a search for the inputs where its score is wrong. We close the section with the metrics whose reporting makes a verifier checkable by someone else.
Section~\ref{sec:conclusion} concludes.

\section{Human Verifiers}
\label{sec:human}

A human verifier asks a person to look at what the policy did and say whether the behavior is good. The four methods in this section differ in what the person sees and in what the person is asked to return. The person sees two trajectories and says which one is better (\S\ref{sec:human:compare}). The person sees one trajectory and gives a score at every time step (\S\ref{sec:human:score}). The person watches a policy while it runs, decides whether to take over, and supplies the correct action (\S\ref{sec:human:intervene}). The fourth method changes the object rather than the question, showing the person a generated rollout and asking whether that rollout preserves the outcome the real execution would have reached (\S\ref{sec:human:source}). All four methods share one limitation, that verdicts are costly to obtain. A human verdict can apply to any task and is the label closest to the intended task, so these verdicts also serve as training data for cheaper verifiers. Each subsection below covers how its labels enter the cheaper verifier's training.

\begin{figure}[!t]
\centering
\includegraphics[width=0.95\linewidth]{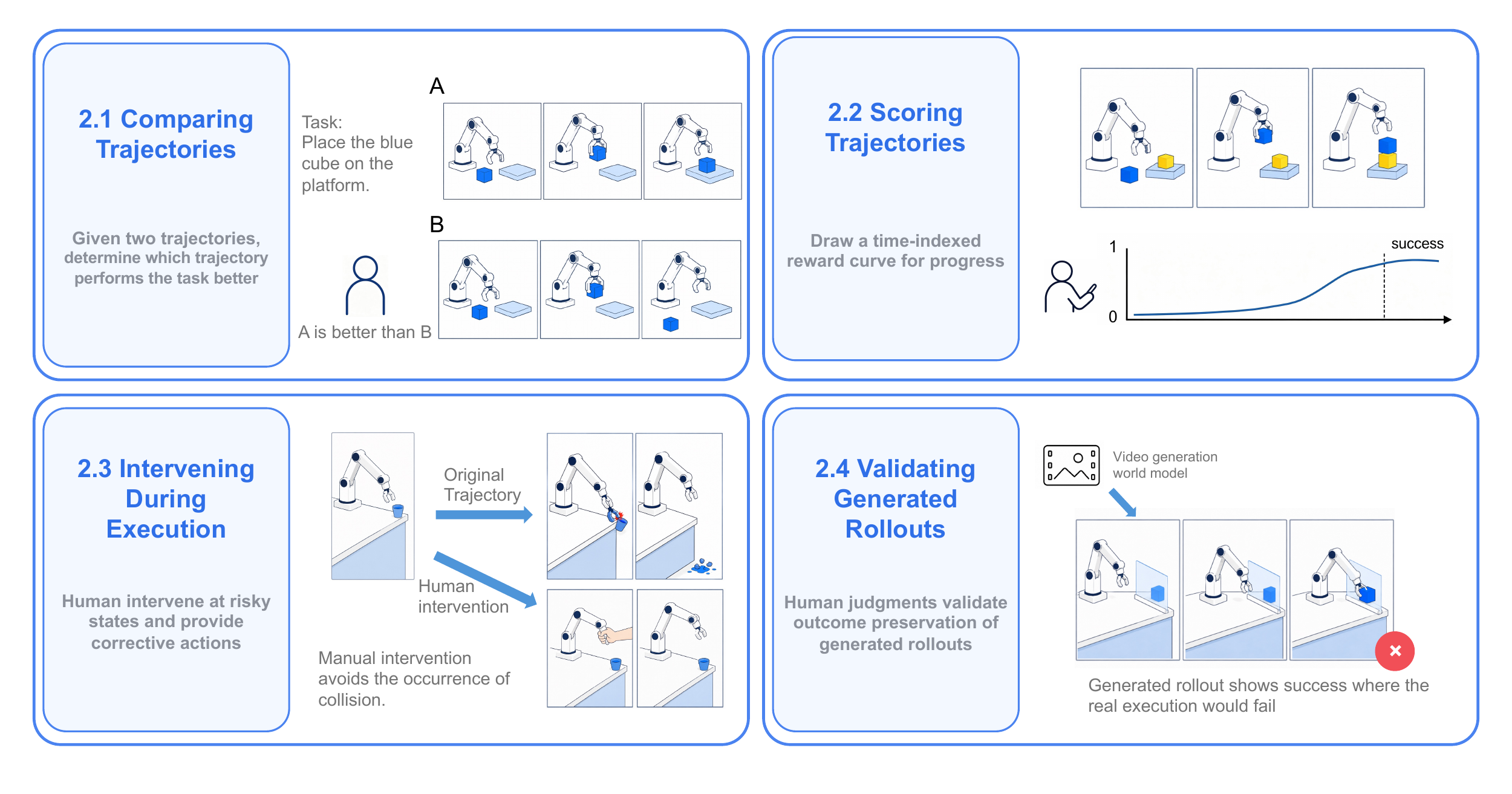}
\caption{Four roles of the human verifier. A person can compare trajectories (\S\ref{sec:human:compare}), assign scalar or time-indexed scores (\S\ref{sec:human:score}), or intervene during execution (\S\ref{sec:human:intervene}). The first three roles judge policy behavior, and the fourth validates generated rollouts and the verifiers applied to them (\S\ref{sec:human:source}).}
\label{fig:human}
\end{figure}

\subsection{Comparing Trajectories}
\label{sec:human:compare}
\label{sec:human:elicit}

A human preference comparison presents two trajectory segments and asks which segment shows better task performance, and the output is therefore a relative judgment. RoboArena uses this judgment directly, conducting double-blind comparisons of fixed policies and aggregating the human preferences into a policy ranking \citep{RoboArena}.

Preference-based learning uses the same interface as supervision for a scalable learned verifier. \citet{ChristianoPrefs} train a reward model from human comparisons and iteratively use the predicted rewards of that model to train a policy model. PEBBLE reduces the required human labels through unsupervised pre-training and by relabelling stored experience whenever the reward model changes \citep{PEBBLE}. B-Pref replaces people with simulated teachers whose controlled irrationalities test the robustness of preference-based reinforcement-learning algorithms \citep{BPref}. Across these methods, the interface remains pairwise, but the judge source moves from direct human judgment to a learned or simulated proxy.

\subsection{Scoring Trajectories}
\label{sec:human:score}

Instead of comparing two trajectories, a person can provide scalar feedback by assigning a numerical score to each time step in a single trajectory. \citet{RewardSketching} call this annotation procedure reward sketching. The annotator draws a curve with values from 0 to 1 over the timeline of a recorded robot trajectory, while the tool displays the video frame corresponding to the current point on the curve, and the curve therefore records perceived progress toward the target task frame by frame. A small set of these curves trains a reward model that then scores every time step of a much larger corpus.

\subsection{Intervening During Execution}
\label{sec:human:intervene}

During execution, a human supervisor can watch the system and take over the controls when the policy model is about to act incorrectly. The intervention methods below split by who decides that the moment to take over has come, a person in the human-gated setting and a learned verifier in the robot-gated setting.

In the human-gated setting, the person decides when autonomous control stops and supplies corrective actions. The intervention serves two purposes. An intervention can prevent an unsafe or unproductive action before that action causes failure, and it provides a corrective training example at a state where the policy model needs help. HG-DAgger studies this interaction in simulated and real-world autonomous driving \citep{HGDagger}, extending the correction-collecting loop of DAgger to a supervisor who chooses the moments \citep{ross2011dagger}. The driver takes over in unsafe states, and the corrective actions become demonstrations for training the policy model. Although the training-time interventions of HG-DAgger remain human-gated, HG-DAgger also calibrates a threshold on the policy's own uncertainty using the states where the driver took over, and the trained policy model can therefore flag comparable states on its own. HACO applies human-gated interaction to simulated autonomous driving, using partial demonstrations to shape a proxy value function that guides policy optimization while discouraging reliance on further interventions \citep{HACO}. Sirius applies human-gated intervention to real-robot manipulation during deployment. The operator takes over at difficult states, and the recorded interventions are weighted by the robot's past reliability at that state before entering the policy update \citep{Sirius}. HIL-SERL keeps the operator in the loop through real-world reinforcement learning, where corrective takeovers enter the replay buffer alongside demonstrations and drive precise manipulation tasks to near-perfect success within a few hours \citep{HIL-SERL}.

Continuous human monitoring is costly, and robot-gated methods therefore automate the intervention decision. In simulated and physical manipulation tasks, ThriftyDAgger uses a switching policy between robot and human control to request help when the current state is unfamiliar or likely to lead to task failure, subject to a human-intervention budget \citep{ThriftyDAgger}. AIM learns a proxy Q-function that approximates the human intervention rule and requests assistance when the learned proxy value indicates the policy model's action deviates from the expert's action, tested in autonomous-driving and grid-world navigation simulations \citep{AIM}. Sirius-Fleet moves the same decision to a deployed fleet, where visual world models predict failure ahead of time and the thresholds are relaxed as the policies improve \citep{liu2024siriusfleet}. In both settings the person still supplies the corrective actions. An intervention verifier must request help early enough to avoid failure but selectively enough to limit human effort.

\subsection{Validating Generated Rollouts}
\label{sec:human:source}
\label{sec:human:instrument}

The first three roles judge the behavior of a real robot. The fourth role judges a generated rollout, produced by a world model or a simulator in place of a real execution. Reading such a rollout takes two judgments, one on the task success of the policy that acted and one on the fidelity of the video the world model produced. The person is asked whether the rollout ends in the same success or failure that the real execution would have reached, and we call this property outcome preservation. GigaWorld-1 asks for both judgments in one number. Annotators rate long-horizon generated rollouts on a four-level World Model as Evaluator Score, which reads task success together with video fidelity \citep{GigaWorld-1}. A fine-tuned vision-language model then reproduces these ratings across a corpus too large for any annotator to watch. VeoEval keeps a person as the judge of each generated rollout. VeoEval runs policies inside a Veo world simulator over nominal, out-of-distribution, and red-teaming conditions, and a person scores the success of each generated rollout \citep{VeoEval}. When outcome preservation fails, a generated rollout shows the task succeeding where the robot would have failed, and the policy then looks better than its real performance. The reference for outcome preservation is the outcome that the real execution would have reached, and with no real rollout available, a person reads the reference directly. A model tuned on the person's judgments then applies that reference at scale.

Across all four roles, the human verdict becomes supervision for a cheaper verifier. Comparisons and scores train reward models, interventions enter the replay buffer as corrective demonstrations, and ratings of generated rollouts train the model that rates the rest of the corpus. The cost of each substitution depends on how far the cheaper verifier's training data is from the data the verifier is later applied to (\S\ref{sec:cross:transfer}).

\section{Rule-Based and Formal Verifiers}
\label{sec:rules}

\begin{figure}[!t]
\centering
\includegraphics[width=0.95\linewidth]{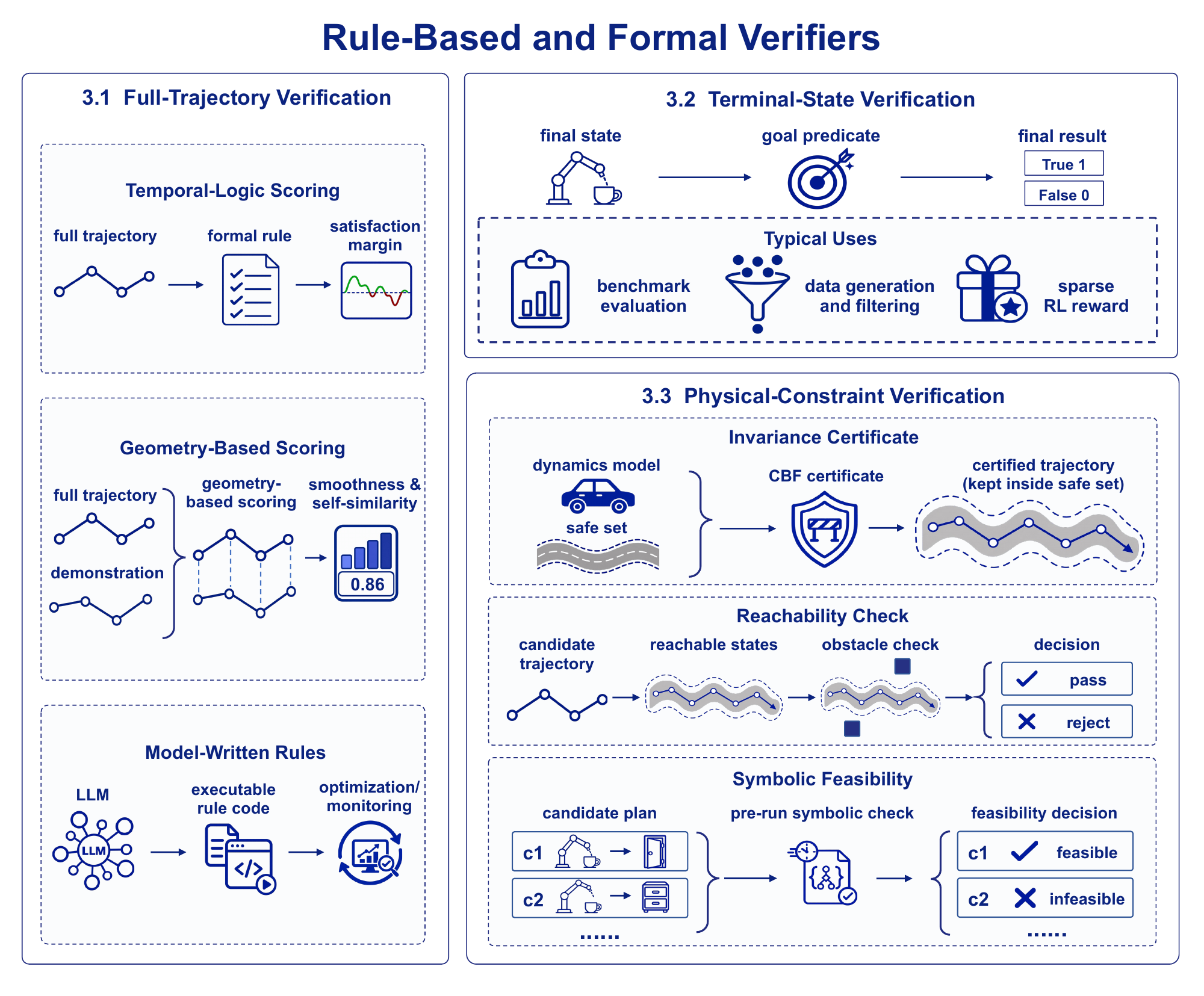}
\caption{Overview of rule-based and formal verifiers. Full-trajectory methods (\S\ref{sec:rules:whole}) score complete rollouts using temporal-logic specifications, trajectory geometry, or model-written code; terminal-state methods (\S\ref{sec:rules:goal}) check whether the final state satisfies the goal; and physical-constraint methods (\S\ref{sec:rules:model}) verify safety through an invariance certificate or a reachability check on each candidate trajectory, and feasibility through a symbolic check before anything moves.}
\label{fig:rules}
\end{figure}

A rule-based or formal verifier applies a criterion written before the run, and therefore returns a verdict without asking a person at inference time.
We group these procedures by what the criterion reads, a full trajectory (\S\ref{sec:rules:whole}), a terminal state (\S\ref{sec:rules:goal}), or a physical constraint (\S\ref{sec:rules:model}). Figure~\ref{fig:rules} shows the three groups, and Table~\ref{tab:rules-index} names the seven methods those groups contain.

\begin{table}[h]
\centering
\footnotesize
\setlength{\tabcolsep}{5pt}
\renewcommand{\arraystretch}{1.2}
\begin{tabular}{@{}>{\raggedright\arraybackslash}p{4.6cm}>{\raggedright\arraybackslash}p{6.2cm}>{\raggedright\arraybackslash}p{5.0cm}@{}}
\toprule
\textbf{Method} & \textbf{Core Characteristic} & \textbf{Representative Work} \\
\midrule
\multicolumn{3}{c}{\textit{Full-trajectory Verification}} \\
\midrule
Temporal-Logic Scoring (\S\ref{sec:rules:whole:spec}) & Quantifies \textbf{\textit{how much}} a trajectory satisfies a formal specification & ConformalSTL \citep{ConformalSTL}, TGPO \citep{TGPO} \\
Geometry-Based Scoring (\S\ref{sec:rules:whole:geometry}) & Returns a \textbf{\textit{smoothness}} score and a \textbf{\textit{self-similarity}} score against a smoothed version of the trajectory itself & RINSE \citep{RINSE} \\
Model-Written Rules (\S\ref{sec:rules:whole:authored}) & LLM \textbf{\textit{writes}} the verification code, ranked by a task metric or only tested for execution & Eureka \citep{Eureka}, Code-as-Monitor \citep{Code-as-Monitor} \\
\midrule
\multicolumn{3}{c}{\textit{Terminal-state Verification}} \\
\midrule
Goal Predicate (\S\ref{sec:rules:goal}) & A \textbf{\textit{true/false}} check on the final state, simulation only & LIBERO \citep{LIBERO}, CALVIN \citep{CALVIN} \\
\midrule
\multicolumn{3}{c}{\textit{Physical-constraint Verification}} \\
\midrule
Invariance Certificate (\S\ref{sec:rules:model}) & A \textbf{\textit{formal proof}} that holds for all possible trajectories & PNCBF \citep{PNCBF} \\
Reachability Check (\S\ref{sec:rules:model}) & A \textbf{\textit{per-trajectory}} check of whether the robot's reachable states intersect an obstacle & RTD \citep{RTD}, ARMOUR \citep{ARMOUR} \\
Symbolic Feasibility (\S\ref{sec:rules:model}) & A \textbf{\textit{pre-run check}} that a plan is physically possible & Text2Motion \citep{Text2Motion} \\
\bottomrule
\end{tabular}
\caption{Seven verification schemes with representative works.}
\label{tab:rules-index}
\end{table}

\subsection{Full-Trajectory Verification}
\label{sec:rules:whole}

Trajectory-level verification methods process full state sequences. Three types appear, hand-written temporal-logic scoring (\S\ref{sec:rules:whole:spec}), geometry-based trajectory scoring (\S\ref{sec:rules:whole:geometry}), and model-written verification code (\S\ref{sec:rules:whole:authored}).

\subsubsection{Temporal-Logic Scoring}
\label{sec:rules:whole:spec}

Temporal-logic scoring is the trajectory-level method appearing most often in the work we cover. The method states the criterion as a formal specification and scores a trajectory by its distance from satisfying that specification. A specification states the required behavior and its timing, in a formal language a program can check a run against. This literature writes specifications in Signal Temporal Logic (STL), and the score is a real number called the satisfaction margin:
\begin{equation}
\operatorname{satisfaction\text{-}margin}\bigl(\mathit{specification},\;\mathit{trajectory}\bigr) \in \mathbb{R}.
\label{eq:robustness}
\end{equation}
The margin is positive when the specification is satisfied and negative when it is violated, and the magnitude of the margin gives how close the trajectory ran to the boundary. If the specification says ``the gripper must close within 0.25m of the table'', closing at 0.1m gives a positive margin and 0.3m a negative margin, and 0.05m scores higher than 0.2m because it leaves more room. This quantity is called robustness in the temporal-logic literature.

The same satisfaction margin applies to a pre-recorded execution \citep{SafeManip}, to a simulated rollout \citep{TGPO}, and to a future trajectory predicted by a trained model \citep{ConformalSTL}. The margin is computed the same way for every rollout source. Practical systems differ in one respect, whether an extra calibration step is added on top of the raw satisfaction margin.

\paragraph{Calibrated predictive runtime verification.}
Predictive runtime verification calibrates the satisfaction margin with conformal prediction \citep{ConformalSTL,DistRobustPRV}.
Conformal prediction is a method that attaches its own margin to the output of any predictor. The method runs the predictor on a held-out set with known outcomes, records the size of the resulting errors, and takes a quantile of those recorded errors as the margin carried by every later prediction. For example, a margin set at the ninety-fifth percentile of the held-out errors gives an interval that contains the true outcome in ninety-five percent of later cases. The bound holds for any predictor and any data distribution, as long as the held-out cases and the cases judged later are drawn from the same pool, so the order of the two sets carries no information about their errors. Drawing both from the same pool is the condition called exchangeability, and \S\ref{sec:write:performative} shows how training a policy against a verifier breaks that condition.

In \citet{ConformalSTL}, the trajectory predictor is treated as a black box and predicts future states from current observations. The calibrated error bounds of that predictor become an interval on the satisfaction margin, and the system can therefore warn that a specification is about to be violated several time steps in advance.

Two extensions carry conformal bounds on the satisfaction margin into settings the original method excludes. First, robust conformal prediction handles a distribution shift between calibration and deployment \citep{RobustConformalSTL}. Second, a distributionally-robust version extends the approach to spatio-temporal specifications \citep{DistRobustPRV}. A related method keeps the calibration and drops the specification. \citet{SLS2} work in the latent space of a world model, where they conformally calibrate the dynamics-prediction error and a learned safety-classifier score. The calibrated bounds build a robust constraint set for model-predictive control, and safety constraints are therefore enforced during closed-loop execution from pixels.

\paragraph{Uncalibrated specification evaluation.}
Specifications can also be used without any calibration layer, either as an offline monitor or as an optimization reward.
SafeManip defines temporal safety properties in Linear Temporal Logic over finite traces (LTLf), a logic for specifying sequences of discrete events over a finite time horizon. SafeManip compiles each property into an automaton and checks pre-computed simulation rollouts against these automata \citep{SafeManip}. The output is a boolean violation with a timestamp and duration. These safety properties read the simulator's state directly, and the same checks on hardware therefore require perception, which SafeManip leaves for future work.

When exact ground-truth states are available, the satisfaction margin can be used directly as an optimization reward, providing a dense learning signal at every time step of policy training. TGPO decomposes an STL formula into subgoals and invariant constraints, and builds a staged dense reward from them, solving the sparsity problem that makes specification-driven learning difficult \citep{TGPO}. Cycle experience replay compiles LTL into an automaton and shapes a dense reward from its accepting cycles \citep{LTL-CER}.

\subsubsection{Geometry-Based Scoring}
\label{sec:rules:whole:geometry}

Geometry-based scoring reads the spatial shape of the trajectory alone, and no task specification has to be written by hand. This method applies wherever a set of recorded demonstrations is available.

RINSE scores each demonstration with two geometric measures. Spectral arc length gives a smoothness score that penalizes high-frequency jitter, and trajectory envelope distance gives a self-similarity score, measuring how far the trajectory deviates from a Bézier-smoothed version of itself. Both scores are reference-free and need no other demonstrations to compare against. RINSE trains on the highest-scoring subset and performs better than training on the full set \citep{RINSE}.

Spectral arc length and envelope distance measure execution quality rather than task success, and a smooth trajectory can therefore still fail the task while a jerky trajectory can succeed. RINSE's scores agree closely with the mixture weights that Re-Mix learns over a heterogeneous corpus (\S\ref{sec:scorers:data}). These two scores therefore serve as a lightweight filter over training data, and a separate check has to determine whether the task was completed.

\subsubsection{Model-Written Rules}
\label{sec:rules:whole:authored}

A model writes the verification logic itself. The generated code is used in two ways, as a training objective and as a runtime check.

\paragraph{Optimization.} A model asked for a reward function produces code, and a search loop keeps the candidates that train a good policy. Eureka runs the loop \citep{Eureka}. Eureka reads the environment code and a task description, writes candidate reward functions in Python, trains a policy with each candidate, ranks the candidates by the environment's own ground-truth task metric evaluated on those training runs, and mutates the best candidate, repeating the cycle. Ranking candidates by the task metric checks the generated code against what the task requires, and the monitoring systems below have no equivalent check. The reward Eureka converges to outperforms human-written rewards on most tasks in a large benchmark. DrEureka adds safety instructions and automatic domain-randomization ranges so the learned reward transfers to real hardware \citep{DrEureka}. Text2Reward and language-to-rewards generate dense reward code from natural language descriptions, using iterative human feedback instead of an evolutionary loop \citep{Text2Reward,L2R}. Video2Reward derives rewards for legged robot behavior from video instead of text \citep{Video2Reward}.

\paragraph{Monitoring.} Code-as-Monitor uses a multimodal model to generate a spatio-temporal constraint verifier, and applies geometric abstractions of the constrained elements to keep the check fast enough for the control loop \citep{Code-as-Monitor}. Code-as-Monitor tests each generated verifier before use, to confirm that the generated code runs. ReKep uses a model to express manipulation goals as relational keypoint constraints, which a solver then optimizes into end-effector motion at control rate \citep{ReKep}. Neither Code-as-Monitor nor ReKep checks the generated constraint against what the task requires.

\subsection{Terminal-State Verification}
\label{sec:rules:goal}

A goal predicate is a simple true/false test on the final state. A goal predicate is a formal criterion like the temporal-logic specifications of \S\ref{sec:rules:whole:spec}, but it reads only the last state and returns a bit rather than a margin. Such a predicate needs the object's pose, which a physics engine reports exactly but perception only estimates, so this method is confined to simulation. The same cheap test scores benchmarks, generates training data, and supplies reinforcement learning rewards.

\paragraph{Benchmark evaluation.}
A goal predicate appears in policy benchmarks as the built-in success check. LIBERO defines the goal of each task as a predicate over simulator state, evaluates that predicate as the episode runs, and reports the fraction of episodes the predicate accepts, across four suites that vary spatial layout, objects, goals, and combinations of all three \citep{LIBERO}. Most of the reinforcement learning and test-time selection results in this survey are measured on LIBERO, and the sparse reward those runs optimize is exactly this success bit. CALVIN chains five language-conditioned subtasks and counts the number completed in order \citep{CALVIN}. VLABench and GenManip split the predicate into stages for the same reason, so that a long-horizon policy earns partial credit for the stages the policy did reach \citep{VLABench,GenManip}. RoboCasa builds 100 kitchen tasks on the same simulator with generated scenes and assets \citep{RoboCasa}. RoboVerse unifies embodiments and environments across simulators \citep{RoboVerse}. ManiSkill provides parallelized simulation \citep{ManiSkill}, and Isaac Lab-Arena runs a whole evaluation the same way, and the number of trials therefore stops being set by how long a robot takes to run one trial \citep{isaaclab2025}. RoboTwin adds strong domain randomization to a bimanual setting \citep{RoboTwin2}. BEHAVIOR-1K extends the task distribution toward everyday activities \citep{BEHAVIOR-1K}. THE COLOSSEUM holds the task fixed instead and reads one predicate under a fixed set of perturbations to appearance, lighting, and background \citep{pumacay2024colosseum}. Each predicate is written once and then determines every result the benchmark produces, and an error in the predicate therefore affects the ranking of policies. A policy benchmark reports the policy's success rate, and the error rate of the check behind that number is reported far less often (\S\ref{sec:meta:agreement}).

\paragraph{Synthetic data generation and filtering.}
A boolean check on exact simulator state costs almost nothing to run, so the same check can generate training data as well as verify it. MimicGen runs this generate-and-verify loop \citep{MimicGen}. MimicGen takes a small number of human demonstrations, breaks each demonstration into object-centric segments, adapts those segments to new object poses to generate thousands of trials, runs the trials, and keeps only the trials that pass the task-level success check. Generation and verification are the same loop, and the predicate makes the loop run automatically. DexMimicGen uses the same approach for bimanual dexterous manipulation \citep{DexMimicGen}. DemoGen targets real hardware instead, where running the generated trials to verify them is expensive, and synthesizes spatially augmented point-cloud observations without running or checking any trial \citep{DemoGen}.

Three systems move the task design itself into the loop. GenSim uses a language model to write simulation task code and expert demonstrations, and keeps a candidate task only if a single-task policy trained on that task achieves a success rate above a threshold \citep{GenSim}. GenSim2 scales this method to articulated objects with a multimodal planner \citep{GenSim2}. RoboGen automates task proposal, scene construction, and supervision generation, and its authors note that verifying the resulting skills at scale remains an open problem \citep{RoboGen}.

\paragraph{Sparse reward for reinforcement learning.}
A goal predicate also serves as a reinforcement learning reward, one bit per episode. SimpleVLA-RL shows this most directly \citep{SimpleVLA-RL}. SimpleVLA-RL runs Group Relative Policy Optimization (GRPO) on a VLA model, uses the simulator's success check as the only reward, and provides the parallel rendering and exploration changes needed to keep training stable. A policy initialized from just one trajectory per task then reaches near-perfect performance. RIPT-VLA achieves a similar result with leave-one-out policy-gradient estimation on the same sparse binary reward, deliberately avoiding shaped rewards, value functions, and reward models \citep{RIPT-VLA}. RLinf-VLA provides the infrastructure for running such training across different algorithms and benchmarks \citep{RLinf-VLA}. ConRFT applies a similar construction to real-world fine-tuning, substituting a binary classifier trained on human-labeled demonstrations for the simulator predicate. ConRFT combines behavior cloning and Q-learning in one objective under a consistency policy, shifting the weight from behavior cloning to Q-learning as training moves from offline to online \citep{ConRFT}.

The predicate needs exact object pose, and the same question on hardware therefore requires a different verifier, where pretrained multimodal models have become the default (\S\ref{sec:scorers}). The predicate also reads only the final state, and a run can satisfy the predicate exactly and still fail the task. A policy trained against such a check learns to aim at exactly the runs that satisfy the predicate without doing the task, a pattern demonstrated so far on LLM verifiers for inductive-reasoning tasks rather than on robot policies \citep{GamingVerifiers}. \citet{Unhackability} define this failure mode and \citet{GoodhartTaxonomy} categorize its variants, and \S\ref{sec:meta:hacking} examines how a robot verifier is measured against this failure mode.

\subsection{Physical-Constraint Verification}
\label{sec:rules:model}

Physical-constraint verification requires that the robot and the surrounding objects stay where they are allowed to be. For example, the arm must not touch the person standing next to the robot, and the cup at the edge of the table must not be pushed off. Three methods appear below, ordered by how many trajectories one verdict covers. First, an invariance certificate takes an analytical model of the system dynamics and tests whether the robot can be kept out of unsafe states. The proof covers every trajectory those dynamics can produce. Second, a reachability check computes the states the robot can reach while following a plan, either one candidate trajectory or a family of plans analyzed in advance. The verdict covers the plans that were analyzed. Third, a symbolic feasibility check takes the preconditions of each action and tests whether a plan is physically possible at all. It answers before any trajectory exists.

\paragraph{An invariance certificate that holds for every trajectory.} The certificate here is a control barrier function \citep{ames2019cbf}, a mathematical tool that takes the dynamics model and keeps the robot inside a safe set, the region of states the robot is allowed to occupy. The barrier function proves that a robot starting inside the region stays inside the region under the closed-loop dynamics, a property called forward invariance, and the guarantee therefore holds for every trajectory those dynamics can produce. \citet{PNCBF} construct such a function from the value function of a nominal policy, and the result is a filter that makes any policy safe. A filter of this kind monitors the control input that a policy proposes and overrides that input when the input would leave the safe set, a construction that also covers predictive filters solving a short constrained optimization over a fixed horizon at every control step \citep{wabersich2021psf,hsu2024safetyfilter}. The theorem applies to the exact value function. A neural network approximating that function carries the same guarantee only when a separate verification step checks the descent condition, and the experiments in \citet{PNCBF} do not perform that step. \citet{ValueIsCBF} identify value functions with control barrier functions directly, and supply the verification step that PNCBF's experiments skip. A theorem there extends the guarantee to an $\epsilon$-optimal approximate value function, and validity and coverage metrics check a learned approximation against that theorem.

\paragraph{A reachability check on each candidate trajectory.} The check here reads one candidate trajectory and tests whether the robot's reachable states intersect an obstacle. The methods below differ in where the candidate comes from, running from a family of trajectories analyzed before the robot runs, to the trajectory a planner intends to execute, to the action a policy proposes while it trains.

Before the robot runs, reachability-based trajectory design (RTD) computes every state the robot can reach while tracking any trajectory in a family described by a few parameters, with the tracking error of the controller included \citep{RTD}. At run time, the sensed obstacles are mapped through these precomputed states into constraints on the trajectory parameters, and the planner optimizes only over parameters that satisfy them, fast enough to fit inside the planning loop. Every trajectory in the family ends by braking to a stop, so when no candidate passes, the robot finishes its previous verified plan and stops. The computed set is conservative, so a candidate that is in fact safe can be rejected. Later methods apply the same computation to a manipulator with uncertain mass and inertia \citep{ARMOUR} and to full vehicle dynamics \citep{REFINE}. PARC instead works backward from the goal, under-approximating the set of initial states and plan parameters from which the robot is guaranteed to reach the goal and avoid the obstacles, tightly enough that goal reaching stays feasible in a narrow gap \citep{PARC}. NeuralPARC extends that computation to a black-box robot by fitting a ReLU network to its trajectory data \citep{NeuralPARC}. \citet{ChouWAFR} compute a reachable set of the same kind when the state is estimated from images, adding a bound on the error of a learned perception module that holds with high probability in a trusted domain near the training data. This family of parameterized plans verified before the robot runs is closely related to formally verified maneuver automata, in which each motion primitive is verified by reachability analysis \citep{HessManeuver}.

Fail-safe motion planning checks the trajectory a planner intends to execute \citep{PekFailSafe}. The planner executes that trajectory only while a second trajectory branching off it can bring the vehicle to states from which the vehicle stays safe for all future time, and a convex optimization finds that second trajectory online in dense urban traffic. When the obstacle is a person, SaRA computes the positions the person can reach within the planning horizon \citep{SaRA}.

A safety layer checks the action a policy proposes while it trains, so the verifier stays in the loop for the whole run. \citet{ThummSafeRL} train a manipulation policy next to a person, in simulation with recorded human motion. The layer allows an action only if the arm can still come to a full stop before the person can enter the arm's range, and brings the arm to that stop otherwise. Because safety-critical collisions no longer end the episode, the layer also improves the learned policy. RTS precomputes reachable sets in the same way as RTD and puts them in the loop, checking the trajectory parameter the agent picks and replacing it with a safe parameter when the check fails \citep{RTS}. BRSL computes the reachable states directly from data, without a model of the robot \citep{BRSL}. Continuous action masking gives the policy the whole set of allowed actions rather than a verdict on a single action, and reports that focusing exploration this way also speeds up training \citep{ActionMasking}. The allowed set is a safe action set in some of its tasks and a task-knowledge constraint in others.

\paragraph{A symbolic feasibility check before anything moves.} Symbolic feasibility checking tests whether the preconditions and affordances (the action possibilities offered by the environment) of a candidate plan can be satisfied \citep{TAMPSurvey}. Such a check catches plans that sound reasonable but are physically impossible, before anything moves. A line of work called task and motion planning puts the check between a symbolic plan and the motions that realize the plan, and returns the plan to the symbolic layer when no motion satisfies it \citep{PlannerInterface}. Text2Motion applies the same test to plans that a language model proposes, scoring a candidate skill sequence by whether the policies for those skills can execute that sequence \citep{Text2Motion}. The coverage of Text2Motion is limited by the symbolic abstraction it uses, and that abstraction is maintained by hand.

\section{Learned and Pretrained Verifiers}
\label{sec:scorers}

\begin{table}[!t]
\centering

\newcolumntype{R}{>{\raggedright\arraybackslash}m{4.1cm}}
\newcolumntype{Q}{>{\centering\arraybackslash}m{3.75cm}}

{\footnotesize
\setlength{\tabcolsep}{3pt}
\setlength{\aboverulesep}{0pt}
\begin{tabular}{@{}c@{\hspace{7pt}}c@{}}
\begin{NiceTabular}{@{}R QQ@{}}[cell-space-top-limit=3pt,cell-space-bottom-limit=3.5pt]
\CodeBefore
  \cellcolor{prcF}{2-2,2-3,4-2}
  \cellcolor{prcE}{3-2}
  \cellcolor{prcC}{3-3}
  \cellcolor{prcB}{4-3}
\Body
\toprule
\textbf{Applications} & \textbf{Trained verifier} & \textbf{Pretrained verifier} \\
\midrule
Evaluating behaviour and policies (\S\ref{sec:scorers:evaluation})
  & \prw{VIP~\citep{VIP}, Robometer~\citep{Robometer}, AHA~\citep{AHA}, RoboFAC~\citep{RoboFAC}, AutoEval~\citep{AutoEval}}
  & \prw{GVL~\citep{GVL}, TOPReward~\citep{TOPReward}, WorldEval~\citep{WorldEval}, WorldGym~\citep{WorldGym}} \\
Inference-time verification and scaling (\S\ref{sec:scorers:testtime})
  & \prw{RoboMonkey~\citep{RoboMonkey}, RoVer~\citep{RoVer}, Foresight~\citep{Foresight}, AHEAD~\citep{AHEAD}}
  & \prn{KnowNo~\citep{KnowNo}, FOREWARN~\citep{FOREWARN}} \\
Feeding verifier feedback into policy learning (\S\ref{sec:scorers:learning})
  & \prw{RECAP~\citep{RECAP}, VLA-RL~\citep{VLA-RL}, Demo-SCORE~\citep{Demo-SCORE}, RoboCat~\citep{RoboCat}, Re-Mix~\citep{Re-Mix}, SCIZOR~\citep{SCIZOR}}
  & \prn{SRPO~\citep{SRPO}} \\
\bottomrule
\end{NiceTabular}
&
\heatbar
\\
\end{tabular}
\par}

\caption{Verifier-based methods are organized along two complementary dimensions. The application role is evaluating behavior and policies, inference-time verification and scaling, or feeding verifier feedback into policy learning. The model source is task-trained or used directly from pretraining. Fill depth indicates the number of unique papers in each cell.}
\label{tab:scorers-split}
\end{table}

A learned or pretrained verifier returns its score from a forward pass of a neural network. Those networks come from two sources of supervision, training on task-specific robot data and pretraining alone, and Table~\ref{tab:scorers-split} counts the papers under each source. The two sources call for different evidence that the verifier is right. A verifier trained on task-specific robot data can be compared with real outcomes on the robot the verifier will be used on, and a verifier that comes from pretraining alone has to transfer across embodiments and task distributions before any such comparison exists (\S\ref{sec:cross:transfer}).

\begin{figure}[!t]
\centering
\includegraphics[width=\linewidth]{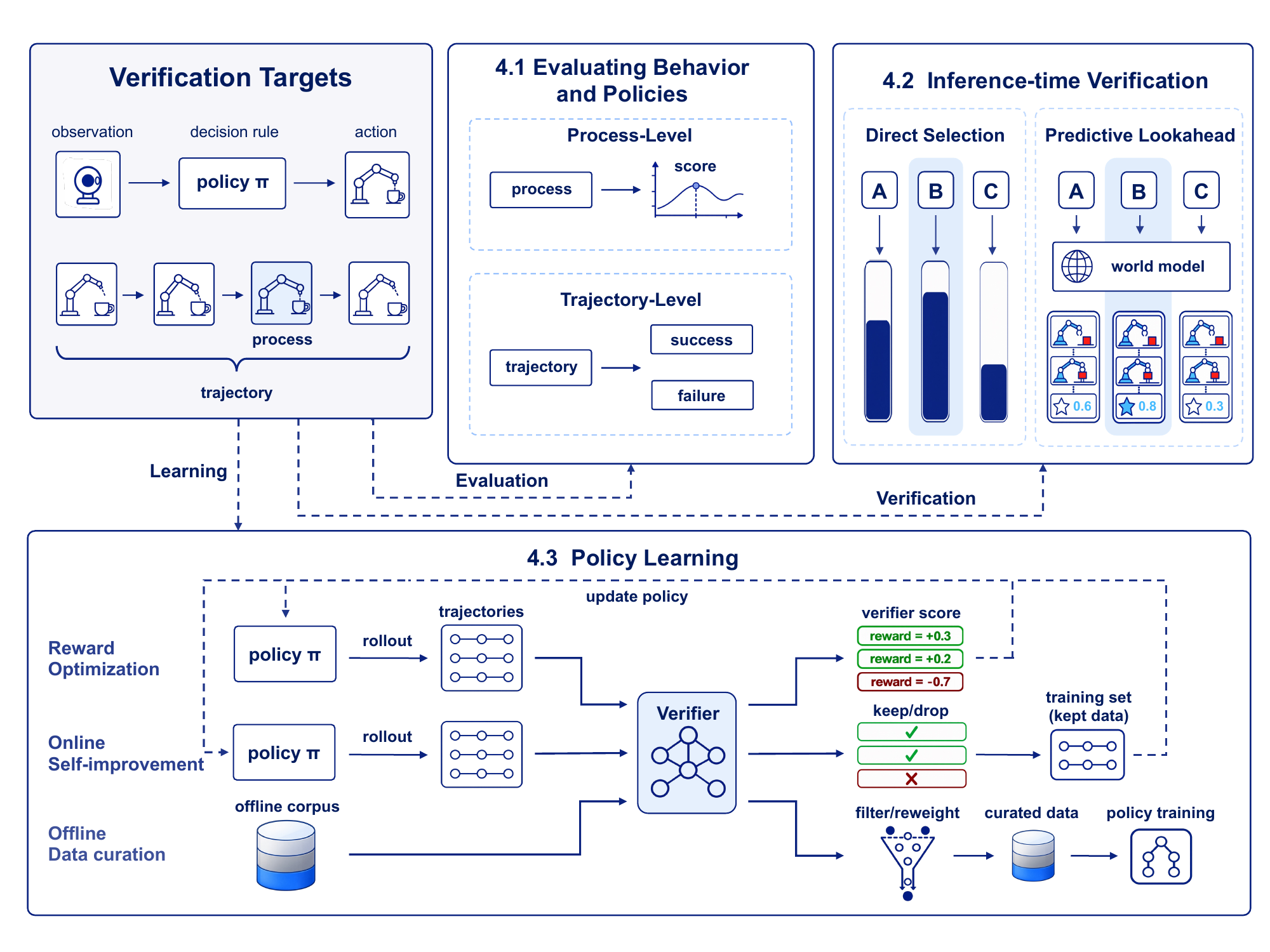}
\caption{Overview of how learned and pretrained verifiers are used in robot learning. They evaluate behavior and policies at process or trajectory level (\S\ref{sec:scorers:evaluation}), guide inference-time selection directly or through predictive lookahead (\S\ref{sec:scorers:testtime}), and provide feedback for policy optimization and data curation (\S\ref{sec:scorers:learning}).}
\label{fig:scorers}
\end{figure}

A learned or pretrained verifier appears in three roles, shown in Figure~\ref{fig:scorers} and set side by side in Table~\ref{tab:scorers-index}. Such a verifier measures the performance of an execution or a policy (\S\ref{sec:scorers:evaluation}), picks among candidate actions, skills, or plans at inference time (\S\ref{sec:scorers:testtime}), and feeds a score back into policy training or into the choice of training data (\S\ref{sec:scorers:learning}).

\begin{table}[h]
\centering
\footnotesize
\begin{tabular}{@{}>{\raggedright\arraybackslash}p{4.0cm}>{\raggedright\arraybackslash}p{2.1cm}>{\raggedright\arraybackslash}p{8.0cm}@{}}
\toprule
\textbf{Application subsection} & \textbf{Axis} & \textbf{Distinguishing feature} \\
\midrule
Evaluating behavior and policies (\S\ref{sec:scorers:evaluation}) & Evaluation & Measuring an execution at process or trajectory level \\
Inference-time verification and scaling (\S\ref{sec:scorers:testtime}) & Decision & Selecting the best action, skill, or plan during inference \\
Feeding verifier feedback into policy learning (\S\ref{sec:scorers:learning}) & Optimization & Using scores to optimize the original policy \\
\bottomrule
\end{tabular}
\caption{The three applications of learned and pretrained verifiers and the distinguishing axes.}
\label{tab:scorers-index}
\end{table}

\subsection{Evaluating Behavior and Policies}
\label{sec:scorers:evaluation}

Models that score robot execution can be organized at two levels, set by the granularity of their outputs. A process-level verifier assigns a temporally localized score or categorical judgment to a state, transition, short clip, trajectory prefix, or bounded subtask. A trajectory-level verifier returns one holistic score or verdict for a completed rollout. Policy evaluations and rankings are then obtained by aggregating these outcomes across repeated rollouts.

\paragraph{Process-level scores within a rollout.}
\label{sec:scorers:dense}

VIP learns an action-free, goal-conditioned value representation from human videos and derives dense, smooth rewards for unseen robotic tasks \citep{VIP}. LIV extends this formulation to goals given as either language or an image, and assigns a dense reward to each frame \citep{LIV}. Robo-Dopamine learns a general-purpose, step-aware process reward model from multi-view inputs \citep{Robo-Dopamine}, while ProcVLM grounds frame-level progress estimates in procedural stages and within-stage visual change \citep{ProcVLM}. Robometer combines frame-level progress supervision with trajectory-comparison preference supervision, and the resulting reward stays available inside a rollout \citep{Robometer}. VLAC outputs a dense progress increment together with a done signal from paired observations and a language goal \citep{VLAC}. Large reward models scale the same dense per-frame reward, training one frame-level reward generator across two dozen data sources and using the frozen generator to refine a policy online in a closed loop \citep{wu2026lrm}. PRM-as-a-Judge turns a dense score of this kind into an evaluation protocol, on the argument that one success bit collapses the execution behind that bit \citep{prm_as_a_judge2026}. PRM-as-a-Judge reads policy execution from trajectory video and reports an outcome, a progress score built on a task-aligned potential, and a diagnosis.

Pretrained VLMs can also provide localized feedback without requiring a separately trained robot reward model. \citet{rocamonde2024vlmrm} establish the construction, using the similarity between a frame and a language description of the goal under a frozen vision-language model as the reward, and RoboCLIP builds the same similarity score from a single demonstration video but assigns that score once, as a sparse reward at the end of the episode, rather than to each frame \citep{sontakke2023roboclip}. GVL presents shuffled trajectory frames to a frozen VLM and autoregressively predicts a task-completion percentage for each frame, producing in-context value estimates at frame level \citep{GVL}. TOPReward converts the likelihood of an affirmative completion token for a video prefix into a dense, instruction-conditioned reward and also supports terminal success detection \citep{TOPReward}. Other methods learn localized categorical judgments. SuccessVQA accepts a single image or short video clip, classifies each clip for success, and only subsequently consolidates the clip predictions into an episode-level decision \citep{SuccessVQA}. AHA, after being instruction-tuned on a large synthetic dataset of failure trajectories, predicts a binary success condition for the current subtask from the keyframes observed up to the subtask and generates a failure explanation when the prediction is negative \citep{AHA}. AHA follows REFLECT, which summarizes a robot's sensory record into a hierarchy and lets a language model query it to explain a failure and propose a correction \citep{liu2023reflect}.

\paragraph{Trajectory-level outcome judgments.}
\label{sec:scorers:success}

RoboFAC takes a complete manipulation video and determines whether the task was successfully completed. On a failed video, RoboFAC localizes the failure to a substage and returns corrective guidance, which the authors show can be fed back to a running VLA policy in real time to improve real-robot task success \citep{RoboFAC}. WorldEval uses a video-language verifier to assign one binary success label to each policy-conditioned generated rollout \citep{WorldEval}. WorldGym passes the generated frame sequence and task instruction to a VLM and returns one rollout-level reward, optionally using a task-specific partial-credit rubric \citep{WorldGym}. AutoEval executes a policy for a fixed trial horizon and applies a learned success classifier to the terminal state, producing one binary outcome for each unattended hardware trial \citep{AutoEval}. RobotArena $\infty$ translates recorded real scenes into simulation and scores the rollouts that a policy produces there with a vision-language model \citep{robotarena_infty2025}. For generated rollouts, GigaWorld-1 presents a complete synchronized multi-view video to a tuned VLM, which outputs one ordinal World Model as Evaluator Score together with supporting rationales \citep{GigaWorld-1}. The World Model as Evaluator Score reads video fidelity together with task success (\S\ref{sec:human:source}).

\subsection{Inference-Time Verification}
\label{sec:scorers:testtime}

At inference time a verifier can choose what the robot does next. The system generates several candidate actions, skills, or short-term plans, scores the candidates with the verifier, and executes the highest-scoring candidate.

\paragraph{Direct candidate selection.}\label{sec:scorers:preexec} RoboMonkey samples and perturbs policy outputs, scores the candidates with a multimodal verifier, and finds that action error falls as an exponentiated power law in the number of samples \citep{RoboMonkey}. RoVer adds an external test-time verifier and improves inference performance with the policy weights left frozen \citep{RoVer}. Three further systems spend more computation at test time, on task-progress scoring, on an adaptive critic over relative actions, and on a general framework for scaling embodied inference \citep{TapSampling,VLA-ATTC,E-TTS}. SayCan evaluates the candidate skills before execution \citep{SayCan}; KnowNo uses conformal calibration to construct a prediction set over candidate actions and requests human assistance when multiple actions remain plausible \citep{KnowNo}. V-GPS uses a value function learned through offline reinforcement learning to re-rank action chunks generated by a frozen generalist policy \citep{V-GPS}.

\paragraph{Predictive lookahead with world models.}\label{sec:scorers:generator} These methods use a world model to look ahead from the current state, either scoring candidate actions by their predicted consequence or monitoring the consequence of the chosen action. AHEAD rolls a motion-aware latent world model forward once to predict future scene tokens rather than candidate actions, and estimates uncertainty from the variance across five sampled rollouts before feeding the predicted tokens to a frozen action decoder \citep{AHEAD}. FOREWARN aligns a latent predictive model with a vision--language model, enabling the system to make semantic judgments about predicted outcomes and guide policy selection accordingly \citep{FOREWARN}. Critics trained on a policy's own successes and failures score the video-model-predicted terminal frames of fine-grained candidate actions \citep{RobotCritics}. Unlike these methods, Foresight does not compare candidates. Foresight reads the latent representations of an action-conditioned world model to monitor the single action block the policy has committed to, detecting potential failures in long-horizon manipulation tasks and conformally calibrating these predictions using trajectory-level labels \citep{Foresight}.

\subsection{Feeding Verifier Feedback into Policy Learning}
\label{sec:scorers:learning}

Verifier feedback can also change the policy itself, by entering the training objective or by deciding which data the policy is trained on.

\paragraph{Direct reward optimization.}\label{sec:scorers:reward} The verifier's output becomes a reward, an advantage, or a preference objective, and the policy parameters move with that output. RECAP trains a value function using demonstrations, on-policy rollouts, and teleoperated interventions, and then extracts a policy through advantage conditioning. At deployment scale, RECAP more than doubles system throughput and roughly halves the failure rate on the most challenging household tasks \citep{RECAP}. GRAPE decomposes tasks into stages, constructs a cost function for each stage, and trains the policy through trajectory-level preference optimization \citep{GRAPE}. RL-VLM-F trains a reward model from preferences provided by a multimodal model \citep{RL-VLM-F}, a construction that Motif introduced by distilling a language model's preferences over event captions into an intrinsic reward \citep{klissarov2024motif}, whereas VLA-RL fine-tunes a multimodal backbone into a process reward model that scores automatically extracted task segments \citep{VLA-RL}. 

Some methods derive optimization signals purely from a world model's latent representations, without decoding a video. SRPO compares states in a pretrained latent world model to construct a progress-shaped reward and can assign credit to useful behavior within failed trajectories \citep{SRPO}. World value models put a value head on a world-model backbone instead of a vision-language backbone, on the argument that a backbone without temporal modeling cannot order states along a task \citep{wang2026wvm}. EVA instead generates a full video from the world model and rewards the policy for predictions from which an inverse dynamics model can recover executable actions \citep{wang2026eva}. VLA-RFT performs reinforcement fine-tuning in a world simulator using verified rewards \citep{VLA-RFT}. Although these methods use world-model latent representations, as predictive lookahead does, their final purpose is to update policy parameters rather than to select the next action to execute.

\paragraph{Self-improvement and rollout filtering.}\label{sec:scorers:curation} A verifier decides which newly generated rollouts re-enter training. That decision closes a generate-filter-train loop, and the policy improves on its own experience. Demo-SCORE learns to distinguish successful from failed rollouts and filters a heterogeneous demonstration set before retraining \citep{Demo-SCORE}. RoboCat fine-tunes on a new task, generates its own interaction experience, and uses a verifier to select the experiences that enter subsequent training \citep{RoboCat}. Self-improving embodied foundation models derive both a reward function and a success detector from steps-to-go prediction, enabling autonomous practice with improved sample efficiency \citep{SelfImprovingEFM}. Human-in-the-loop online rejection sampling combines corrective interventions with a reward-aware data filter on contact-rich tasks \citep{Hi-ORS}, while real-time verification scores intermediate results as embodied reasoning is generated and uses the verified reasoning data for subsequent skill learning \citep{RealTimeVerif}.

\paragraph{Data curation and demonstration scoring.}\label{sec:scorers:data} A curation method scores, filters, or reweights a corpus recorded before the current policy existed. The property these methods score is the training value of a demonstration, the contribution that demonstration makes to the policy trained on it. The three groups below differ in the evidence they read for that contribution.

The first group reads the demonstration by itself. DemInf scores a trajectory with a mutual-information estimate over its observations and actions, a quantity falling when the recorded actions look arbitrary given the robot's observations \citep{DemInf}. ILEED attributes quality to the person who recorded the trajectory and estimates each demonstrator's expertise \citep{ILEED}. DWBC trains a discriminator to separate expert demonstrations from the rest and weights each sample by the discriminator's output \citep{DWBC}, and DemoDICE keeps matching the expert distribution while still drawing on imperfect demonstrations \citep{DemoDICE}. 

The second group reads the corpus around the demonstration. Re-Mix learns mixture weights over Open X-Embodiment \citep{OXE}, a corpus aggregated across embodiments and protocols. Re-Mix optimizes the worst-performing domain, so the weight a sub-dataset receives does not follow from its size, and the result outperforms uniform mixtures and hand-specified compositions \citep{Re-Mix}. SCIZOR combines a self-supervised progress estimator that flags suboptimal sub-trajectories with a deduplication module that drops segments repeating what the corpus already holds, and ablations show the progress estimator contributes more of the two \citep{SCIZOR}. 

The third group reads the policy the data produces. DataMIL fits datamodels that estimate the contribution of one training example to downstream performance, without running a new robot rollout \citep{DataMIL}. Curation needs this third quantity, and this quantity is also the most expensive to obtain, since it is defined through a training run. All three groups change the corpus. A policy can instead keep the corpus and take the demonstration's training value as an input, training on the weaker demonstrations with a label that marks them and conditioning at test time on the behavior to follow \citep{RECAP}. $\pi_{0.7}$ does the same with manually annotated quality and mistake labels in place of a trained value \citep{pi07_2026}. Curation and conditioning need the same score, and only curation has to commit to that score before training begins. Two results bound the reach of any such score. The first distinguishes action divergence from transition diversity and finds that state diversity helps only in some settings \citep{DataQualityIL}, and the second finds that corpus composition matters more as the corpus grows \citep{DataScalingLaws}.

\section{Model-Intrinsic Verifiers}
\label{sec:intrinsic}

\begin{figure}[!t]
      \centering
      \includegraphics[
          width=0.98\linewidth,
          trim=0 0 0 0,
          clip
      ]{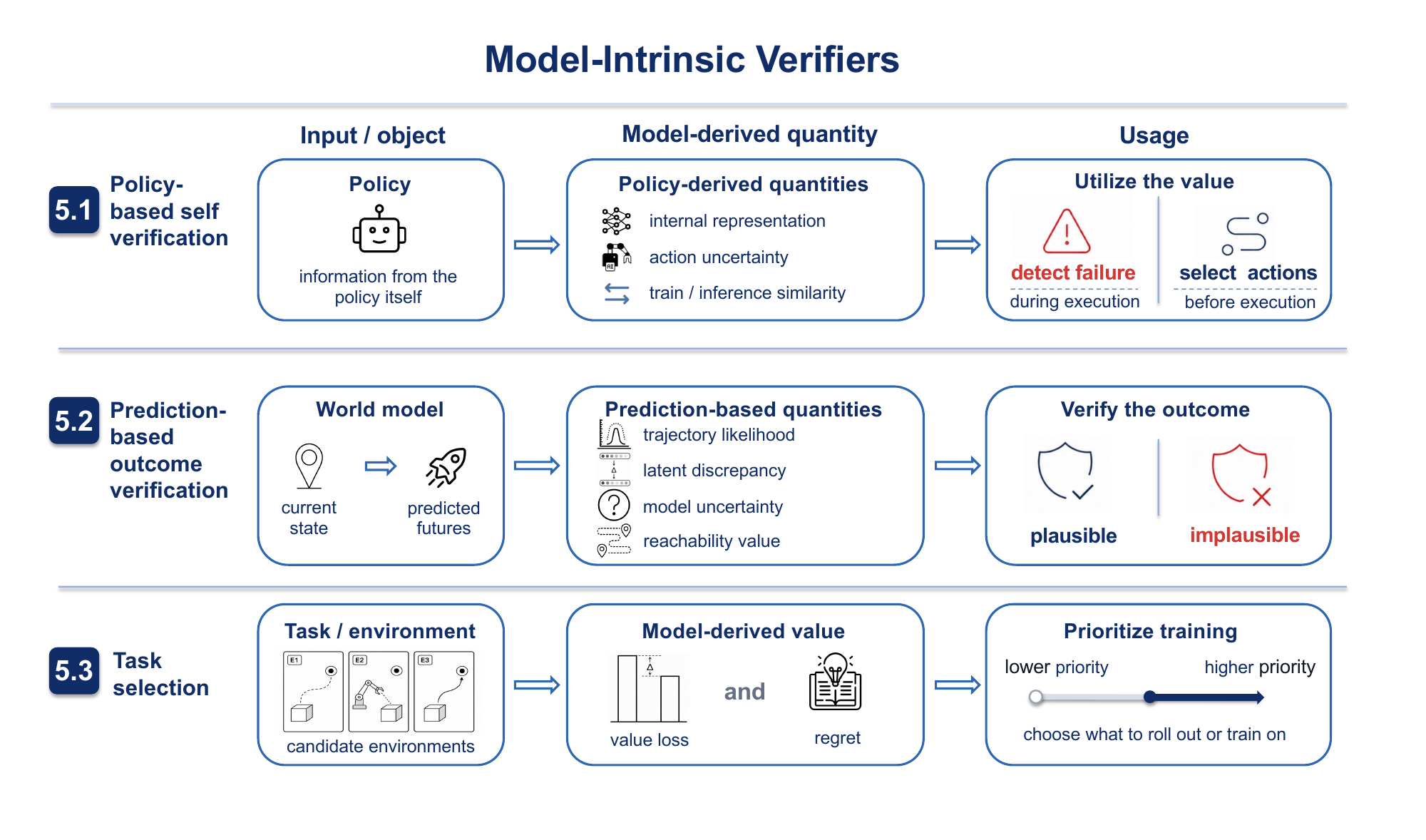}
      \caption{
          Model-intrinsic verifiers use information from a robot's own policy or world model to evaluate behavior and guide task selection. Policy-based self-verification detects failures and selects candidates, prediction-based outcome verification judges a candidate by the outcome the world model predicts, and task selection prioritizes the next training environment, all without an independent external verifier.
      }
      \label{fig:intrinsic}
  \end{figure}

A model-intrinsic verifier reads a quantity that the robot's own policy or learned world model already computes, and we call that quantity a model-intrinsic score.
\emph{Policy-based self-verification} reads the policy, either to detect a failure while an execution runs or to choose among candidate actions before one is executed (\S\ref{sec:intrinsic:policy}).
\emph{Prediction-based outcome verification} reads a learned world model and judges a candidate by the outcome that the world model predicts for that candidate (\S\ref{sec:intrinsic:generator}).
The same scores can then rank a whole task instead of a single action, and decide which environment the policy trains on next (\S\ref{sec:intrinsic:environment}).
Figure~\ref{fig:intrinsic} shows the three subsections.

\subsection{Policy-Based Self-Verification}
\label{sec:intrinsic:policy}

A robot policy produces more than its final action. The policy's internal representations, the uncertainty or inconsistency in its action predictions, and the compatibility of the current observation-action trajectory with the training distribution can all provide evidence on whether the policy's own behavior is reliable.
Policy-based self-verification is used at two different stages. \emph{Detection} evaluates an ongoing execution and raises an alarm when the behavior begins to deviate from patterns associated with normal operation, often before failure becomes externally apparent. \emph{Selection}, in contrast, evaluates multiple candidate actions or trajectories before execution and chooses the candidate that the policy considers most reliable, without physically trying each alternative.

\paragraph{Verifying behavior during execution.} The detection methods below differ in the internal information they read and in the way that information becomes a failure prediction.
SAFE provides a representative example \citep{SAFE}.
SAFE trains a lightweight predictor on the internal representations of a vision-language-action policy, motivated by the observation that successful and failed executions exhibit separable structure in the policy's latent space, with failure-related patterns transferring across tasks.
At runtime, the predictor maps the history of these representations to a scalar failure score, which is compared with a time-varying threshold calibrated using functional conformal prediction.
When calibration and test rollouts follow the same data distribution, this calibration bounds the probability of falsely flagging a successful rollout; however, the guarantee can fail under cross-task distribution shift, as calibration is performed on seen tasks while evaluation may involve unseen tasks.

Other methods also read the policy's own internal signals to judge whether the current execution is reliable, and these methods differ in the signal they read and in how the verifier is built.
Sentinel monitors both action instability and task progress. Temporal inconsistencies in the policy's action predictions are used to detect erratic behavior, while a separate VLM checks whether the task is still making progress \citep{Sentinel}.
FIPER avoids failure examples by combining out-of-distribution signals from the policy's observation embeddings with uncertainty in generated action chunks \citep{RoemerFailPred},
and FAILDetect similarly learns only from successful rollouts, calibrating a time-varying conformal-prediction threshold band on their scores with a statistical guarantee on the false-positive rate \citep{FAILDetect}.
Tri-Info instead targets cross-policy transfer, scoring entropy and mutual information over states and actions, and the resulting measure therefore does not depend on a particular embedding architecture \citep{Tri-Info}.
HideAndSeek reduces the supervision requirement in another way, using only trajectory-level outcomes together with contrastive learning to identify which steps in a failed rollout carry failure signals \citep{HideAndSeek}.

\paragraph{Verifying candidates before execution.} MG-Select scores candidate action chunks by the change in their token probabilities when state and language inputs are masked, then executes the most input-dependent candidate \citep{MG-Select}. The base form of MG-Select needs no extra training and no external model, though the strongest results use a finetuned variant, and the method has only been evaluated on autoregressive policies with discrete action tokens.
TACO instead selects candidates that are most familiar in the policy's representation space, using a separately trained pseudo-count estimator while keeping the policy itself frozen \citep{TACO}. 

The central limitation of policy-based self-verification is that these scores measure whether a behavior looks reliable or familiar to the policy itself, not whether the task succeeded. A familiar-looking failure may therefore still receive a high score.

\subsection{Prediction-Based Outcome Verification}\label{sec:intrinsic:generator}%

A learned world model predicts how the environment may evolve. Those predictions can be used to evaluate whether an outcome is plausible, expected, or still achievable.
Four model-intrinsic scores support such a judgment. \emph{Trajectory likelihood} measures how ordinary the trajectory looks to a video model, \emph{latent discrepancy} compares the predicted latent with the observed latent, \emph{model uncertainty} reads the disagreement inside an ensemble, and \emph{reachability value} says whether failure can still be avoided.

\paragraph{Trajectory likelihood.} VIPER scores a trajectory by its likelihood under a video prediction model \citep{VIPER}.
An autoregressive transformer is trained on expert video, and the prediction likelihoods become an action-free reward.
The resulting agent reaches expert-level control across simulated control suites and manipulation tasks, and also demonstrates transfer across different arm--task combinations.
Diffusion Reward instead uses the conditional entropy of a video diffusion model, based on the observation that conditioning on expert trajectories reduces generative diversity \citep{DiffusionReward}.
A limitation of likelihood-based verification is that high generative likelihood does not necessarily correspond to semantically correct behavior under distribution shift.

\paragraph{Latent discrepancy.}
A latent-discrepancy verifier encodes the observed outcome and compares the resulting latent with the predicted latent.
\citet{VJEPAPhysics} use this discrepancy between predicted and observed latents to detect physical implausibility zero-shot, without training an additional detector.
\citet{VJEPAPhysics} find that prediction in latent space performs substantially better than pixel-space prediction and than multimodal language models on this task.
Within robotics, Foresight reads predicted latents of the same kind from an action-conditioned world model and trains a failure detector on those latents (\S\ref{sec:scorers:generator}).

\paragraph{Model uncertainty.} In model-based offline reinforcement learning, an imagined rollout is penalized in the regions where the learned dynamics model is uncertain.
MOPO penalizes uncertain predictions from an ensemble of learned dynamics models \citep{MOPO}. MOReL instead identifies unknown state--action pairs when an ensemble of dynamics models makes inconsistent predictions \citep{MOReL}.
LOMPO extends this approach to the latent space of a trained model, allowing model uncertainty to be estimated from image observations \citep{LOMPO}.
AHEAD reads the same ensemble disagreement inside a manipulation policy, taking the variance across five sampled latent rollouts as a per-step uncertainty and halting the lookahead once the variance exceeds a threshold \citep{AHEAD}. \citet{ward2026foundational} calibrate the latent uncertainty of a compact action-conditioned world model and use that uncertainty to flag failures while the execution runs, and the detector therefore needs no failure examples of its own. Tested on a diffusion policy and on a real-world bimanual cable-manipulation dataset, a 570K-parameter version of the detector beats the next-best learned baseline by 3.8 points of failure-detection rate.

\paragraph{Reachability value.} A reachability value states whether failure can still be avoided from the current state under the predicted dynamics. Like the candidate scoring of \S\ref{sec:intrinsic:policy}, a reachability value judges an action before the action is executed. The judgment depends on the consequences the world model predicts rather than on how familiar the action looks to the policy.

\citet{LatentSafetyFilters} implement this idea with a generative world model. A classifier first identifies observations that correspond to failure, while the world model predicts how different actions change future states. Hamilton--Jacobi reachability \citep{bansal2017hj} then computes a value over these predicted dynamics, identifying states from which no available action can prevent a future failure. The resulting safety filter can therefore reject an unsafe action and switch to a safety-preserving action before the failure occurs. LS3 is an earlier, related construction from pixels \citep{LS3}. It learns a safe set from the states visited by earlier successful runs and constrains a model-predictive controller to plans that end inside that set.

Follow-up work studies the behavior of this type of verification under partial observability, identifying errors from both state estimation and future prediction and proposing more conservative mitigation strategies \citep{LatentSafetyPO}. Three later methods change the quantity the filter is defined over. A parameterized latent constraint lets the unsafe region be set at run time rather than fixed when the filter is trained \citep{agrawal2025anysafe}. An uncertainty term added to the value treats a state the model has not seen as a state to avoid \citep{seo2025uncertainty}. A learned control barrier function in place of the reachability value makes the filter intervene smoothly rather than switching \citep{nakamura2025latentcbf}. \citet{LatentCertificates} give sufficient conditions under which a barrier function learned in a latent space keeps its guarantee on the original system, stated as an approximate conjugacy between the latent dynamics and the true dynamics. Gameplay Filters run the same forward reasoning in a physics simulator, evaluating safety by simulating interactions between a virtual adversary and the robot's fallback policy \citep{GameplayFilters}.

\subsection{Beyond Action Verification: Task Selection}
\label{sec:intrinsic:environment}
\label{sec:intrinsic:guarantee}

A model-intrinsic score can also be read about a whole environment instead of a single action, and the score then decides which environment the policy trains on next.
ACCEL scores each environment by the positive value loss of the policy running in it \citep{ACCEL}. The value loss is how much the policy's own value function underestimates the return the policy actually received. The environments with the highest score are kept and edited further, and training therefore moves toward the environments the policy still misjudges.
The quantity ACCEL's score approximates is regret, which PAIRED introduced as the difference between the return a second agent achieves in an environment and the return the policy under training achieves \citep{PAIRED}. Measuring regret as this difference needs two agents run in the same environment, and the value loss instead obtains a proxy for regret from the single policy that is already running.
Neither method was evaluated on a robot. PAIRED runs on grid-world maze navigation and a MuJoCo hopper domain, and ACCEL adds lava grids and a continuous-control walker.
In each case the score decides the next environment to roll out, and says nothing about whether a completed behavior succeeded.

Across the three subsections of Section~\ref{sec:intrinsic}, the model-intrinsic score comes from a model the robot is already running rather than from a separate verifier built to check that model. Taking the score from the model itself makes these scores the cheapest verifiers in this survey, and it also means the model gives a high score to a failure the model does not recognize (\S\ref{sec:write}).

\section{Validating the Verifier}
\label{sec:meta}
\phantomsection\label{sec:cross}

A verifier's own error is worth measuring, because a verifier has four uses and a single verification error reaches all four uses. A verifier selects which demonstrations enter training, ranks one policy or one action above another, supplies the reward during post-training, and decides at run time whether an action may execute, for example by switching control to a stronger policy once the verifier flags a coming failure \citep{aegis2026}. One error therefore reaches the training data, the reported ranking, the reward, and the executed action together, and in each of the four uses that error looks like a correct result. Table~\ref{tab:consumption} lists the same four uses, ordered by whether the score is searched against. Curation and ranking select among candidates produced without access to the score, and an error there averages out over the corpus. A training reward and a runtime gate are applied to candidates produced by a search for the inputs where the score is wrong, and one rare wrong-scoring region is then enough to make the score unusable.

\begin{table}[!t]
\centering
\footnotesize
\setlength{\tabcolsep}{5pt}
\begin{tabular}{@{}>{\raggedright\arraybackslash}p{2.5cm}>{\raggedright\arraybackslash}p{3.6cm}>{\raggedright\arraybackslash}p{4.0cm}>{\raggedright\arraybackslash}p{5.2cm}@{}}
\toprule
\textbf{Use} & \textbf{How candidates arise} & \textbf{Requirement on the verifier} & \textbf{Cost of error} \\
\midrule
Curation & fixed corpus, verifier unseen & average accuracy, split by error type & a false positive trains on failure \\ \addlinespace[3pt]
Ranking and off-policy evaluation & non-adversarial sampling & average accuracy, calibration & the wrong checkpoint ships \\ \addlinespace[3pt]
Training reward & closed loop, actively searched & no exploitable region & the policy satisfies the verifier without doing the task \\ \addlinespace[3pt]
Runtime gating & closed loop, millisecond budget & no exploitable region, plus latency & hardware damage \\
\bottomrule
\end{tabular}
\caption{The four uses, ordered by whether the score is searched against. The top two select among non-adversarial candidates. The bottom two expose the score to a search for the regions where the score is wrong. Splitting by error type means counting false positives separately from false negatives, since curation incurs a different cost for each.}
\label{tab:consumption}
\end{table}

Throughout this section a \emph{score} is the output a verifier returns about a behavior, and a \emph{metric} is a quantity we compute about the verifier itself. Three metrics of a verifier's own error appear in the literature. The first is agreement with a fixed reference (\S\ref{sec:meta:agreement}). The second is the policy that results from training on the verifier (\S\ref{sec:meta:downstream}). The third is the verifier's behavior on candidates constructed to score highly without succeeding (\S\ref{sec:meta:hacking}). We then list the metrics whose reporting makes a verifier checkable by someone else (\S\ref{sec:meta:report}).

\subsection{Agreement on a Fixed Set of Rollouts}
\label{sec:meta:agreement}
\phantomsection\label{sec:meta:benchmarks}
\phantomsection\label{sec:meta:chain}
\phantomsection\label{sec:cross:samplesize}
\phantomsection\label{sec:cross:transfer}

An agreement benchmark fixes the rollouts, fixes a reference judgment for each rollout, and reports how often the verifier agrees with the reference. We first name the benchmarks built this way. An agreement rate then omits three things, how a disagreement splits between the verifier and the reference, the confidence interval around the rate, and the kind of rollout the rate was measured on. Table~\ref{tab:resources} closes the subsection with an audit of the resources the field relies on today.

The reward benchmark released with TOPReward runs this benchmark design on 130 tasks across four robot platforms, scoring a candidate reward model by its agreement with human judgment \citep{TOPReward}. Three other benchmarks vary one element of the same benchmark design. The trajectory corpus behind Robometer is far larger, and that corpus was built for training rather than for auditing \citep{Robometer}. RoboReward was built to diagnose verifier error, and among these four benchmarks RoboReward reports the lowest agreement at the time of writing \citep{RoboReward}. OpenGVL takes the true temporal order of the frames as its reference in place of a human label, and tests progress prediction on both robot and human embodiments, so OpenGVL measures whether a verifier still works on a new embodiment \citep{OpenGVL}. The same benchmark design has also been applied to safety. There the reference states whether the behavior is dangerous rather than whether the task succeeded. ASIMOV renders situations taken from real injury reports into text scenarios and videos, then asks a frontier vision-language model to judge those scenarios directly, scoring how well the model recognizes physical danger and decides to intervene \citep{ASIMOV}. A benchmark that reports task success and constraint satisfaction separately yields a number of the same kind \citep{SafeVLABench}. Some rollouts pass the success predicate and break a stated safety property, and the fraction of rollouts doing so is the false positive rate of the predicate. At present, the benchmarks that check a safety property separately from the success predicate agree on one finding: policies that complete more tasks are not safer. Provael carries the same discipline into tooling. It scores a red-teaming attack against a fixed unsafe-region predicate over simulator state, and prints that predicate's own false-positive rate beside every attack-success rate it reports; on the one policy measured so far, the predicate accepts two of fifty rollouts run with no attack applied \citep{provael2026}.

The first of the three things an agreement rate omits is how a disagreement splits between the verifier and the reference. A person is almost always the reference, and the error rate of that person is reported far less often than the agreement rate itself. The opposite case, where the verifier is wrong and the person is right, is also reported rarely. \citet{RoboReward} find that frontier multimodal models used as rewards produce costly false positives and false negatives on tasks that a person judges without difficulty, including calling a failed drawer-opening a success. Such false positives and false negatives belong to the verifier being scored rather than to the human reference the verifier is scored against, so a low reference-error rate would still leave a hidden error on the verifier's side.

The second is the confidence interval around the rate. An agreement rate is reported as a single number, and the width of that interval sets how much evidence the rate provides. The rate itself is the fraction of trials the verifier judged correctly, over samples running from a hundred clips to a few thousand trajectories. Four designs on the policy side each supply one thing that a single number omits. First, blind randomized trials allocate the rollout budget in advance and report the size of that budget \citep{LBM}. Second, a test on the distribution of time to success reports the whole distribution in place of its mean \citep{PhAIL}. Third, bounds for a behavior-cloning policy state the task distribution the bounds hold over \citep{vincent2024generalizable}. Fourth, sequential designs let each next trial be chosen by the data collected so far, rather than fixing the schedule beforehand \citep{STEP,N-SCORE,ActiveEval}. The same machinery is standard outside robotics, in the interval estimators recommended for reinforcement learning comparisons run on too few seeds \citep{agarwal2021precipice}, and in the tests that stay valid when the person running them chooses the moment to stop \citep{ramdas2023safeanytime}.

The third is the kind of rollout. An agreement rate holds for the rollouts the rate was computed on and not for rollouts of another kind. GE-Sim 2.0 makes the difference visible by scoring one verifier twice, on rollouts that its world model generated and on recorded video of the same tasks \citep{GESim2}. A verifier's error rate therefore depends on the kind of input as much as on the verifier itself, and an accuracy reported without naming the input omits the input, which is one of the two things that set the error rate. GE-Sim 2.0 attributes the resulting eight-point accuracy gap to artifacts that the world model itself introduces rather than to the difficulty of the task. The same question arises between simulation and hardware, where SureSim pairs a small number of real trials with simulated trials and shifts the simulated estimate by the average difference \citep{ImperfectSim}, and between embodiments, the difference that OpenGVL measures. Sim-to-real evaluation reads the sim-to-real gap directly, as the agreement between a ranking obtained in simulation and a ranking obtained on the robot, a metric that \citet{yang2025simtorealeval} propose but have not yet measured. RoboChallenge runs a shared real-robot fleet that executes a submitted policy and returns its score, and the hardware ranking is therefore available for comparison \citep{robochallenge2025}. MolmoSpaces reports the sim-to-real correspondence per task, while RoboLab reports the correspondence in aggregate across task suites tagged by the competency that each task requires \citep{kim2026molmospaces,yang2026robolab}.

\citet{wmeval_position2026} argue that a world model should be judged by the decisions made with it rather than by the fidelity of what it generates. The last column of Table~\ref{tab:resources} applies this standard to every resource in this survey. Two of its rows are world-model evaluations whose verifier error was never measured on generated rollouts.

\begin{table}[t]
\centering
\footnotesize
\setlength{\tabcolsep}{4pt}
\resizebox{\linewidth}{!}{%
\begin{tabular}{@{}llll cc@{}}
\toprule
\textbf{Resource} & \textbf{Object scored} & \textbf{Scale reported} & \textbf{Verifier} & \textbf{Serves} & \makecell{\textbf{Verifier error measured on}\\\textbf{its deployment distribution}} \\
\midrule
\multicolumn{6}{@{}l}{\textit{Verifier-quality resources}}\\
ManiRewardBench \citep{TOPReward} & trajectory & 130 tasks, 4 platforms & human & curation, ranking & \cmark \\
RBM-1M \citep{Robometer} & trajectory & $>$1M trajectories & human + auto & curation, ranking & \cmark \\
RoboReward \citep{RoboReward} & trajectory & 45k train, 2.8k verified test & human & curation, ranking & \cmark \\
RoboFAC \citep{RoboFAC} & trajectory & 9{,}440 traj., 78{,}623 QA & human & curation & \cmark \\
OpenGVL \citep{OpenGVL} & trajectory & robot + human embodiments & temporal order & curation & \cmark \\
\midrule
\multicolumn{6}{@{}l}{\textit{Policy-evaluation substrates}}\\
SIMPLER \citep{SIMPLER} & policy & 2 robot suites & simulator rule & ranking & \cmark \\
AutoEval \citep{AutoEval} & policy & 24\,h unattended & learned classifier & ranking & \cmark \\
RoboArena \citep{RoboArena} & policy & 600+ pairwise episodes & human & ranking & \cmark \\
PhAIL \citep{PhAIL} & policy & 4 VLAs, time-to-success CDF & rule + human anchor & ranking & \cmark \\
WorldEval \citep{WorldEval} & policy & 4 policies ranked & frontier VLM & ranking & \xmark \\
WorldGym \citep{WorldGym} & policy & across sizes, versions & VLM & ranking & \xmark \\
WMBench \citep{GigaWorld-1} & world model & 7 models, 324k rollouts & human + VLM & ranking & \cmark \\
GE-Sim 2.0 \citep{GESim2} & policy & 6 tasks, 20 real-robot episodes each & learned verifier & ranking, reward & \cmark \\
\bottomrule
\end{tabular}}
\caption{An audit of the resources used to establish that a verifier works. \textbf{Serves} names the use of the score from Table~\ref{tab:consumption} that the resource supplies evidence for, and the last column states whether the verifier was evaluated on the kind of data the verifier is applied to.}
\label{tab:resources}
\end{table}

\subsection{Measuring a Verifier by the Policy It Trains}
\label{sec:meta:downstream}

A verifier is built to train a better policy or to filter a better corpus. The gain in the resulting policy, rather than agreement with a reference judgment, is the quantity that actually matters. This gain is defined through a training run, and it is therefore expensive to obtain and correspondingly rare in the literature.

An agreement rate and the gain in the trained policy diverge, and the mechanism is training itself. A verifier can agree with a person on ordinary rollouts and still fail on the rollouts that a training run drives the policy toward, because training moves the policy toward exactly those rollouts. Controlled measurements of the divergence between agreement and downstream gain exist on the language side, and we give one in \S\ref{sec:meta:hacking}. A robot verifier is optimized against in the same way, and the mechanism therefore applies on the robot side, where the corresponding measurement has yet to be run.

Two methods in this survey already compute the gain in the trained policy. A datamodel estimates the contribution of one demonstration to downstream performance \citep{DataMIL}, and a self-improvement loop lets a verifier decide which of a policy's own rollouts re-enter training and then measures the result on the retrained policy \citep{Demo-SCORE,RoboCat}. Both methods define verifier quality by the performance of the retrained policy. The agreement rate and the downstream gain are reported separately and rarely for the same verifier, and the two numbers diverge most where a verifier is used as a training reward.

\subsection{Measuring a Verifier Under Optimization}
\label{sec:meta:hacking}
\phantomsection\label{sec:write}
\phantomsection\label{sec:write:exploit}

An agreement rate and a gain in the trained policy are both measured on a verifier that was never searched against. A verifier used as a training reward or a runtime gate has to meet one more requirement, that no region of inputs makes the score wrong in a way that a search can reach. Reward hacking is the failure of that requirement. Some inputs do make the score wrong, such as a run that satisfies a final-state predicate exactly and still fails the task (\S\ref{sec:rules:goal}). A policy optimized against the score is driven toward such wrong-scoring inputs, since the score is highest there. Both measurements were taken on candidates produced without such a search, and both numbers therefore overstate the value of the verifier to a policy trained on it.

Reward hacking is well studied on the language side. A scaling law measures how far a policy can be optimized against a reward model before the true objective starts to fall \citep{RMOveroptimization}, a token-space attack finds inputs that a top-ranked reward model scores highest and should reject \citep{TokenSpaceAttack}, and a survey documents the practice \citep{RewardModelingSurvey}. Known fixes include information-theoretic bottlenecks and non-negative reward constructions \citep{InfoRM,BNRM}. A weak verifier produces reward gains that an independent panel does not confirm, and the gap widens as training continues \citep{RubricHacking}.

In robotics the same question, how far a policy can be optimized against a learned verifier before its score stops tracking the task, is largely undeveloped. Robustness to hacking is reported by the authors of each reward model, as a property of that model. A shared benchmark that holds the candidates fixed and scores the exploitability of each verifier has yet to appear. \citet{RewardAsAgent} argue that the limit on reinforcement learning in embodied world models is unreliable verification rather than exploration. That work builds an agentic reward framework to resist hacking under distribution shift. One method reports on the embodied side of exploitability, letting a policy, a world model that simulates the policy, and a reward model that scores the result improve each other over several rounds \citep{guo2026vlaw}. That method runs the co-adaptation described here as the intended procedure rather than as a failure.

Optimization has two consequences for a verifier. The reward gain the verifier reports becomes an overestimate, and the calibration the verifier was released with stops holding.

\paragraph{Verifiers held out from training.}
One protocol would measure a robot verifier's exploitability without waiting for a shared benchmark. Score the rollouts twice, once with the verifier that the policy was trained on and once with a panel of verifiers the policy never saw. A policy that genuinely improved passes both. A policy that only learned to satisfy its training verifier passes the first scoring and fails the panel. The panel on a robot is a set of trajectory verifiers, and the protocol runs at each checkpoint. Take the rollouts that the training verifier passed, hand those rollouts to the panel, and report the fraction the panel rejects. The simulators and the training setups this protocol needs are published \citep{SimuScene,SimpleVLA-RL}.

\paragraph{A calibration that optimization invalidates.}
\phantomsection\label{sec:write:performative}A conformal margin is a fixed number added to a verifier's score so that the true outcome falls inside the resulting interval a stated fraction of the time. The margin comes from a held-out set \citep{KnowNo,SAFE}. A held-out set is a batch of rollouts whose outcomes are known and which the verifier did not see during its construction. The margin provides its stated coverage as long as the rollouts judged later come from the same pool as the held-out set. The requirement that later rollouts come from the same pool is exchangeability (\S\ref{sec:rules:whole:spec}). Filtering rollouts by the verifier's own scores draws the later rollouts from a different pool, and the margin then no longer provides its stated coverage \citep{ConformalCovShift,ConformalBeyondExch}.

Training a policy against a verifier changes the data that the verifier will later receive \citep{Performative}. The verifier moves the distribution toward the region where the verifier is wrong, and a correction made once therefore stops holding after the next round of training. Conformal failure detection therefore holds for a frozen policy \citep{Foresight} and weakens for a policy trained against the detector \citep{SAFE}, and the same exchangeability argument implies SureSim's bias correction would need to be recomputed once a retrained policy changes the deployment distribution, though the original work evaluates only a single frozen policy \citep{ImperfectSim}. Online conformal methods answer this by recalibrating continuously \citep{AdaptiveConformal}, and the guarantee such methods provide is asymptotic rather than finite-sample.

\subsection{Metrics for Validating a Verifier}
\label{sec:meta:report}
\phantomsection\label{sec:write:checklist}

We provide nine metrics for a verifier, and reporting the nine makes verifier claims comparable across papers. Guidance of this kind exists for evaluating robot policies empirically \citep{EmpiricalScience,LBM,PhAIL}, and the verifier benchmarks of \S\ref{sec:meta:agreement} put a verifier under test. Such a benchmark scores one verifier on one fixed set of rollouts against one fixed reference, and reports a single agreement rate. The nine metrics ask for what the rate omits, from the conditions it was measured under to what becomes of the verifier once a policy is optimized against it. Table~\ref{tab:checklist} splits the nine by whether the score is used to select or optimized against. One of the nine is computed from two separate scorings, and we give the computation below.

\paragraph{Proxy gain that transfers.} This number measures the fraction of a reward gain that is real. Score the first and the last checkpoint under the training verifier and call the difference $\Delta_{\mathrm{proxy}}$. Score the same two checkpoints under a panel that the policy never trained against and call the difference $\Delta_{\mathrm{panel}}$. Report the ratio $\Delta_{\mathrm{panel}}/\Delta_{\mathrm{proxy}}$ together with both differences, since the ratio becomes unstable when $\Delta_{\mathrm{proxy}}$ is small. One means the panel confirms the whole gain, zero means the panel confirms none of the gain, and a negative number means the panel scores the policy lower while the training reward rises. The two differences behind the ratio are separately available in simulation.

\begin{table}[t]
\centering
\footnotesize
\setlength{\tabcolsep}{4pt}
\begin{tabular}{@{}>{\raggedright\arraybackslash}p{4.3cm} >{\raggedright\arraybackslash}p{5.8cm} >{\raggedright\arraybackslash}p{5.9cm}@{}}
\toprule
\textbf{Metric} & \textbf{How it is computed} & \textbf{What it settles} \\
\midrule
\multicolumn{3}{@{}l}{\emph{For a score used to select: curation and ranking}} \\[2pt]
Independent rollout count & count per condition, for every success rate given & the confidence interval can be rebuilt from the count (\S\ref{sec:cross:samplesize}) \\
Error rate by type & false positives and false negatives separately, with the label count behind each & a false positive trains on a failure, a false negative discards usable data \\
Rollout type & teleoperated, scripted, or sampled from the policy being judged & a verifier scoring generated rollouts needs its error measured on generated rollouts (\S\ref{sec:cross:transfer}) \\
Agreement with human labels & agreement rate with its label count, plus agreement between two annotators & how far the verifier sits from the labels, and the labels from each other \\
Calibration & score minus observed success rate of the rollouts in that score bin & required of any continuous score consumed as a reward \\
Cross-embodiment transfer & error on a held-out embodiment minus error on the training embodiments & whether the verifier still works on a new robot \citep{OpenGVL} \\
\midrule
\multicolumn{3}{@{}l}{\emph{For a score optimized against: training reward and runtime gating}} \\[2pt]
Proxy gain that transfers & $\Delta_{\mathrm{panel}}/\Delta_{\mathrm{proxy}}$, with both gains reported & separates a real gain from a gain that only satisfies the training verifier \citep{RubricHacking} \\
Verifier error over training & error on initial-checkpoint and on final-checkpoint rollouts, labeled alike & the increase in verifier error produced by the optimization \\
False positives under search & candidates found by searching the verifier for inputs the verifier wrongly accepts \citep{fuzzing_verifiers2026} & the region a policy trained on this verifier moves toward \\
\bottomrule
\end{tabular}
\caption{Nine metrics that make a verifier claim comparable across papers. Most of the upper block asks for numbers the authors have already measured. The lower block requires running a search against the verifier, which robotics has yet to do for a learned verifier.}
\label{tab:checklist}
\end{table}

\section{Conclusion}
\label{sec:conclusion}

A verifier in robot learning decides what a policy is trained on and what the policy's reported performance means.
We surveyed roughly 150 verifiers and grouped them by who supplies the judgment, into human verifiers, rule-based and formal verifiers, learned and pretrained verifiers, and model-intrinsic verifiers.
We compared the four families on their \emph{availability} and their \emph{credibility}.
Across the four families, \emph{availability} and \emph{credibility} move in opposite directions.
As the source of judgment moves closer to the model being evaluated, a verification signal becomes cheaper to obtain, arrives earlier, and can be queried more often, while the evidence that the signal reflects actual task performance becomes harder to assemble.

The four families occupy different points on the availability-credibility trade-off.
Human verifiers give the most direct reference to the intent of the task, and their verdicts are costly to obtain, so human verifiers appear mostly as the small sample that validates the other three families (\S\ref{sec:human}).
Rule-based and formal verifiers are inexpensive and repeatable once the required state information is available, and they reach the strongest guarantees in this survey, though only where the predicate, the state estimate, and the dynamics assumptions represent the task (\S\ref{sec:rules}).
Learned and pretrained verifiers are inexpensive to query and dense across many tasks and trajectories, and the error rate of such a model depends on the data the model was trained and validated on (\S\ref{sec:scorers}).
Model-intrinsic verifiers cost the least, because the policy or the predictive model already computes the model-intrinsic score. The model-intrinsic score describes the model rather than the task, and its relation to actual task performance is therefore indirect (\S\ref{sec:intrinsic}).
Learned and pretrained verifiers are the largest group in this survey, and model-intrinsic verifiers cost the least to obtain.
Those are exactly the two families whose \emph{credibility} is least often established.

Establishing \emph{credibility} is a separate task from building a verifier.
We surveyed two ways of establishing credibility.
One is a chain of comparisons ending at a set of human labels, and the error of those labels is rarely reported alongside the claim the labels support (\S\ref{sec:meta:chain}).
The other is a benchmark that puts the verifier under test and reports the verifier's own error rate (\S\ref{sec:meta:benchmarks}).
The demand on a verifier depends on how its score is used. Curation and ranking tolerate an average error rate, and a training reward and a runtime gate demand more, because the candidates they score come from a search for the inputs where the score is wrong. Every agreement rate and every downstream gain that this survey collected was measured on non-adversarial candidates (\S\ref{sec:meta}).
\emph{Availability} and \emph{credibility} are meant as coordinates for the verifiers still to be built.
Naming who supplies the judgment fixes how much a verifier costs, how often the verifier can be queried, and how much the verifier's score can prove.
Today, there is \textbf{no free checker}.
Reaching high \emph{availability} and high \emph{credibility} together is the main future work in verification for robot learning.

\phantomsection
\section*{Use of Large Language Models}
\label{sec:bench}

Parts of the text in this survey were drafted with a large language model. Every claim, citation, and count was checked by the authors against the cited papers.

\bibliography{refs}
\bibliographystyle{unsrtnat}

\end{document}